\documentclass[lettersize,journal]{IEEEtran}
\usepackage{amsmath,amsfonts}
\usepackage{algorithmic}
\usepackage{algorithm}
\usepackage{array}
\usepackage[caption=false,font=normalsize,labelfont=sf,textfont=sf]{subfig}
\usepackage{textcomp}
\usepackage{stfloats}
\usepackage{url}
\usepackage{verbatim}
\usepackage{graphicx}
\usepackage{cite}
\usepackage{bm}
\usepackage{colortbl}
\usepackage{multirow}
\usepackage{tabularx}
\usepackage{booktabs}
\usepackage{hyperref}
\usepackage{bbding}
\usepackage{xcolor}
\usepackage{colortbl}  
\usepackage{threeparttable} 
\usepackage{makecell}
\usepackage{supertabular}
\usepackage{algorithm}
\usepackage{pifont}
\usepackage{upgreek}
\usepackage{arydshln}
\usepackage{makecell}
\usepackage{soul}
\usepackage{xcolor}
\usepackage{tablefootnote}
\usepackage{arydshln}
\hypersetup{
	colorlinks=true,
	linkcolor=cyan,
	filecolor=blue,      
	urlcolor=red,
	citecolor=green,
}

\begin{document}
\title{NWPU-MOC: A Benchmark for Fine-grained Multi-category Object Counting in Aerial Images}
	\author{Junyu Gao,~\IEEEmembership{Member,~IEEE}, Liangliang Zhao,  and Xuelong Li,~\IEEEmembership{Fellow,~IEEE}

}

% The paper headers
\markboth{Journal of \LaTeX\ Class Files,~Vol.~14, No.~8, August~2021}%
{Shell \MakeLowercase{\textit{et al.}}: A Sample Article Using IEEEtran.cls for IEEE Journals}

\maketitle

\begin{abstract} 
Object counting is a hot topic in computer vision, which aims to estimate the number of objects in a given image. However, most methods only count objects of a single category for an image, which cannot be applied to scenes that need to count objects with multiple categories simultaneously, especially in aerial scenes. To this end, this paper introduces a \underline{M}ulti-category \underline{O}bject \underline{C}ounting (MOC) task to estimate the numbers of different objects (cars, buildings, ships, \emph{etc.}) in an aerial image. Considering the absence of a dataset for this task, a large-scale Dataset (NWPU-MOC) is collected, consisting of 3,416 scenes with a resolution of 1024 $\times$ 1024 pixels, and well-annotated using 14 fine-grained object categories. Besides, each scene contains RGB and Near Infrared (NIR) images, of which the NIR spectrum can provide richer characterization information compared with only the RGB spectrum. Based on NWPU-MOC, the paper presents a multi-spectrum, multi-category object counting framework, which employs a dual-attention module to fuse the features of RGB and NIR and subsequently regress multi-channel density maps corresponding to each object category. In addition, to modeling the dependency between different channels in the density map with each object category, a spatial contrast loss is designed as a penalty for overlapping predictions at the same spatial position. Experimental results demonstrate that the proposed method achieves state-of-the-art performance compared with some mainstream counting algorithms. The dataset, code and models are publicly available at \url{https://github.com/lyongo/NWPU-MOC}.
\end{abstract}


\section{Introduction}
\IEEEPARstart{O}{bject} counting in aerial scenes aims to estimate the number of objects with a specific category in given images, which can be applied to urban planning \cite{rathore2016urban,551936, 5764519}, environmental monitoring and mapping \cite{pekel2016high, 4358862, 8444430}, disaster detection \cite{zhang2022edge, 1704990, DSRL2023}, and other practical applications \cite{pcc, gao2023dacc, zhao2023naskernel}. Over the past years, benefiting from the development of deep learning and neural networks, some object (pedestrians, cars, boats, \emph{etc.}) counting algorithms designed for remote sensing scenes have been proposed. 
Specifically, Bahmanyar \textit{et al.} \cite{bahmanyar2019mrcnet} introduces a crowd-counting dataset captured from the UAV view. They also propose a multi-resolution network to estimate the pedestrian count. Gao \textit{et al.} \cite{gao2020counting} construct an RSOC dataset for remote sensing object counting, containing four subsets: buildings, small vehicles, large vehicles, and boats.
Based on the dataset, Gao \textit{et al.} propose the PSGCNet \cite{PSGCNet}, which addresses challenges such as scale variations and complex backgrounds in remote sensing scenes by extracting and fusing multi-scale and global feature information.

\begin{figure*}[!ht]
	\centering
 	\includegraphics[scale=0.58]{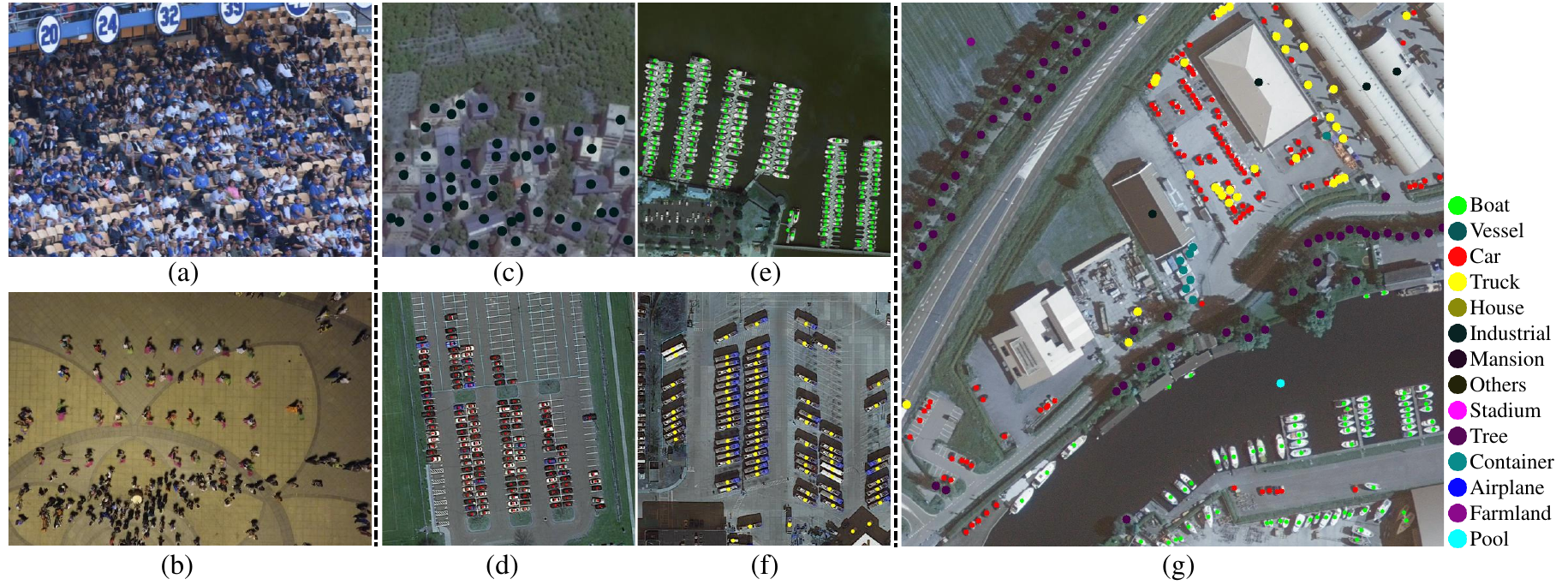}
	\caption{Sample images from some object counting datasets. (a) is a sample from a crowd-counting dataset (NWPU-Crowd) \cite{nwpucrowd}. (b) is a sample from a crowd-counting dataset in the UAV view (DroneCrowd)  \cite{wen2021detection}. (c)-(f) are samples from a remote-sensing object counting dataset (RSOC) \cite{gao2020counting}. (g) is a sample from the NWPU-MOC dataset constructed by this paper. Different from other counting datasets, NWPU-MOC provides annotations for multiple object categories within a single image.}
	\label{fig:1}
\end{figure*}

A few studies proposed counting objects of multiple categories within a single image. Go \mbox{\textit{et al.}} \mbox{\cite{9506384}} established an object counting dataset containing five distinct categories of cranes. The MOCSE13 \mbox{ \cite{MOCSE13}} dataset, created by Liu \textit{et al.}, encompasses a variety of vegetables and fruits and was built using the Unity engine.
As shown in Fig. \ref{fig:1}, compared with other scenes (Fig. \ref{fig:1} (a), (b)), aerial and remote sensing images (Fig. \ref{fig:1} (c)-(g)) often cover a wider spatial range and contain ground objects with multiple categories. Thus, simultaneously counting objects with multiple categories in an aerial image is a more challenging and practical task.

In this paper, we aim to take a step towards this practical goal and introduce a Multi-category Object Counting (MOC) task. To address the task, a large-scale dataset (NWPU-MOC) is constructed, which consists of 3,416 aerial scenes and is annotated with center points of object instances for 14 fine-grained categories.
Table \ref{tabel:1} shows a comparison of some popular object counting datasets. In contrast with other datasets, the NWPU dataset is annotated for multiple categories and contains both RGB and NIR images with a resolution of 1024 $\times$ 1024 pixels.
Due to factors such as vegetation, lighting variations, and weather conditions, the visibility of objects in RGB aerial images can be affected, resulting in inaccurate predictions and difficulties in object recognition. Near-infrared (NIR) light, with longer wavelengths (typically ranging from 0.76 to 0.90 $\upmu$m in aerial NIR images), can easily penetrate the atmosphere. Therefore, the inclusion of NIR images in the NWPU-MOC dataset has the potential to alleviate occlusion issues in aerial scenes.

Compared with specific-category object counting, MOC not only involves counting objects but also requires distinguishing between the different categories of objects. Based on the NWPU-MOC dataset, a multi-spectral, multi-category object counting framework is proposed that uses a density map-based approach. To alleviate occlusion issues in aerial scenes, the proposed framework takes RGB and NIR images as inputs and introduces a dual-attention module to fuse RGB and NIR features. 
In the density map prediction stage, the proposed framework regresses a density map with multi-channel corresponding to each object category for multi-category object counting. Due to the shared feature space, there exists mutual interference between features among the regressed multi-channel density maps, resulting in overlapping predictions.
Therefore, in order to model the mapping relationships between channels of the density map and object categories, we design a spatial contrast loss. It is jointly optimized with a counting loss function to reduce overlapping predictions between channels in the predicted density maps.
On the NWPU-MOC dataset, we reproduce some single-category object counting methods and propose a new evaluation metric to evaluate the performance of counting algorithms for the MOC task.

The main contributions of this paper are the following:
\begin{itemize}
     \item{Construct a large-scale dataset for multi-category object counting task, which includes 3,416 aerial images with both RGB and NIR spectral.}

  \item{Propose a multi-spectrum, multi-category object counting framework, which fuses RGB and NIR images and regresses multi-channel density maps to count objects.}

    \item{Present a spatial contrast loss for modeling the mapping relationship between different channels in the density map with each object category.}
 
	\item{Design a new evaluation metric to evaluate the performance of counting algorithms for MOC task.}
\end{itemize}

The remaining organization of the paper is as follows. Section II reviews some of the methods and datasets that exist in the field of object counting. Section III describes the proposed dataset in detail. Section IV first introduces the proposed multi-category object counting task and proposes the corresponding methodology and evaluation metrics. Section V illustrates the experimental results of the proposed method on the NWPU-MOC dataset and analyses the effectiveness of the proposed method. Finally, we conclude the whole paper in Section VI.

\begin{figure*}[!ht]
	\centering
	\includegraphics[scale=0.32]{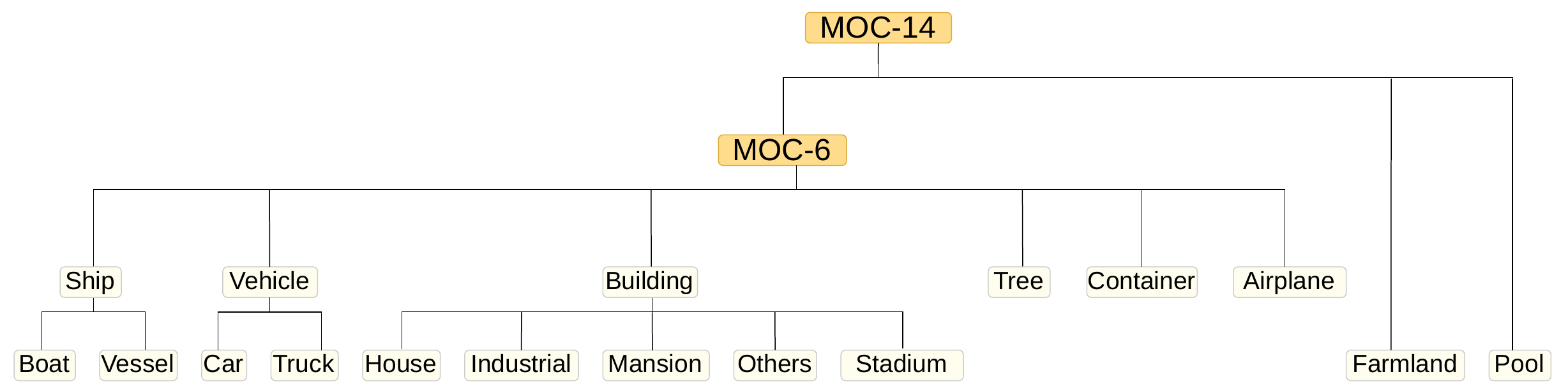}
	\caption{Tree diagram about the full 14 categories (MOC-14) and the grouping of MOC-6 in the NWPU-MOC dataset (Farmlands and Pools are considered negative samples in MOC-6).}
	\label{fig:4}
\end{figure*}

\begin{table*}[!htb]
	%\tiny
	\caption{BRIEF STATISTICAL INFORMATION OF THE PROPOSED NWPU-MOC AND OTHER COUNTING DATASETS.}
	%\vspace{-0.3cm}
	\begin{center}
		\renewcommand{\arraystretch}{1.0}	
		\setlength\tabcolsep{5pt}
		\resizebox{\textwidth}{!}{
			
			\begin{tabular}{c|c|c|c|c|c|c|c|c|c|c}
				
				\Xhline{1.2pt}
				\multicolumn{1}{l}{\multirow {2}{*}{Dataset}}&\multicolumn{1}{|c}{\multirow {2}{*}{Categories}}&\multicolumn{1}{|c}{\multirow {2}{*}{Platform}}&\multicolumn{1}{|c}{\multirow {2}{*}{Multi-spectral }} &\multicolumn{1}{|c}{\multirow {2}{*}{Images}} &\multicolumn{1}{|c}{\multirow {2}{*}{Training/test}} &\multicolumn{1}{|c}{\multirow{2}{*}{Average Resolution}} &\multicolumn{1}{|c}{\multirow {2}{*}{Annotation Format}}
				&\multicolumn{3}{|c}{Count Statistics}\\
				%\hline
				\cline{9-11}
				\multicolumn{1}{c}{}&\multicolumn{1}{|c}{}  &\multicolumn{1}{|c}{} &\multicolumn{1}{|c}{}  &\multicolumn{1}{|c}{} &\multicolumn{1}{|c}{} &\multicolumn{1}{|c}{} &\multicolumn{1}{|c}{} &\multicolumn{1}{|c}{Total}  &Average &Max\\
				\hline
				\multicolumn{1}{l|}{Olive trees~\cite{salami2019fly}}&Tree &UAV &\multicolumn{1}{c|}{\multirow {9}{*}{RGB}} &10 &-- &4000$\times$3000 &circle &1251  &125.1 &143\\
				% \hline
				\multicolumn{1}{l|}{COWC~\cite{mundhenk2016large}}&Car &Aerial & &-- &-- &low &center point &32,716 &-- &-- \\
				% \hline
				\multicolumn{1}{l|}{CARPK~\cite{hsieh2017drone}}& Car&Drone & &1448 &989/459 &1280$\times$720 &bounding box &89,777 &62 &188 \\
				% \hline
				\multicolumn{1}{l|}{DLR-ACD~\cite{bahmanyar2019mrcnet}}&Crowd &Aerial & &33 &19/14 &3619$\times$5226 &center point &226,291 &6857 &24,368 \\
				% \hline
				%			\hline
				\multicolumn{1}{l|}{\multirow{4}{*}{RSOC~\cite{gao2020counting}}}
				&\multicolumn{1}{c|}{Building} &Satellite & &2468 &1205/1263 &512$\times$512 &center point &76,215 &30.88 &142 \\
				%\hline
				&\multicolumn{1}{c|}{Small-vehicle} &Satellite & &280 &222/58 &2473 $\times$ 2339 &oriented bounding box &148,838 &531.56 &8531 \\
				%\hline
				&\multicolumn{1}{c|}{Large-vehicle} &Satellite& &172 &108/64 &1552 $\times$ 1573 &oriented bounding box &16,594 &96.48 &1336\\
				%\hline
				&\multicolumn{1}{c|}{Ship} &Satellite& &137 &97/40 &2558 $\times$ 2668 &oriented bounding box &44,892 &327.68 &1661 \\
				\hdashline

				\multicolumn{1}{l|}{\multirow{1}{*}{KR-GRUIDAE~\mbox{\cite{9506384}}}}
				&\multicolumn{1}{c|}{\makecell[c]{RCC, RCCJ, WNC, WNCJ, AA}} &Ground camera & RGB & 1423 & 2391/1025 &1554$\times$2326 &center point &30,779& 21.63 &299 \\
				\hdashline
				% \hline
				
				\multicolumn{1}{l|}{\multirow{1}{*}{MOCSE13~\cite{MOCSE13}}}
				&\multicolumn{1}{c|}{\makecell[c]{Banana, Avocado \& Apple, Banana, Avocado \& Artichoke,\\  Banana, Orange \& Apple, Carrot, Pear \& Melon}} &Unity &RGB & 120 &72/48 &512$\times$512 & bounding box& -& -& - \\
				\hdashline
				
				\multicolumn{1}{l|}{\multirow{1}{*}{NWPU-MOC(Ours)}}
				&\multicolumn{1}{c|}{\makecell[c]{Boat,Vessel,Car,Truck,House,Mansion,Stadium,\\Tree,Container,Airplane,Farmland,Pool,Others}} &Aerial &RGB\&NIR &3416 &2391/1025 &1024$\times$1024 &center point &383,195&112.18 &3582 \\
				\Xhline{1.2pt}
		\end{tabular}}
	\end{center}
    \vspace{-0.5cm}
	\label{tabel:1}
\end{table*}

\section{Existing Datasets, and Methodologies: an Overview}
In this section, we review some previous research in the field of object counting from two aspects: 1) existing datasets for object counting, and 2) the methods for object counting.

\subsection{Object Counting Datasets}

\subsubsection{Nature Scenes}
Object counting in natural scenes usually focuses on crowd counting. The crowd counting datasets \cite{chan2008privacy, zhang2016single, sindagi2019pushing, wang2021pixel, nwpucrowd, sindagi2020jhu, idrees2018composition, li2022video} primarily contain annotations for pedestrians in dense scenes. The UCSD dataset \cite{chan2008privacy} consists of 2000 image sequences and is collected from outdoor sidewalks with surveillance views at the University of California, San Diego.
ShanghaiTech Part B \cite{zhang2016single} is a high-quality crowd dataset consisting of 782 representative images, which were constructed by Zhang \textit{et al.} and collected from urban congested scenes in Shanghai.
To avoid labor-intensive manual annotations and provide diverse scenes. GCC \cite{wang2021pixel} is constructed by Wang \textit{et al.}, which is a large-scale synthetic crowd dataset by simulating surveillance camera views in the computer game. Sindagi \textit{et al.} \cite{sindagi2019pushing} construct a head-labeled crowd counting dataset, JHU-CROWD, containing 4250 images. Moreover, Wang \textit{et al.} construct a large-scale dataset called NWPU-Crowd, which includes various crowded scenes collected from the internet. These above-mentioned datasets annotate crowd in natural scenes and made significant contributions to the advancement for crowd counting. Only a few works have focused on counting multiple objects within a single image.  As listed in Table \ref{tabel:1}, the KR-GRUIDAE dataset, created by Go \textit{et al.} \cite{9506384}, is based on images captured by ecologists and annotated for four categories of cranes: Red-Crowned Crane (RCC), Red-Crowned Crane Juvenile (RCCJ), White-Naped Crane (WNC), White-Naped Crane Juvenile (WNCJ), and Anser Albifrons (AA) all within the same image. Liu \textit{et al.} \cite{MOCSE13} synthesized the MOCSE13 dataset using the Unity engine, comprising images of 13 different fruits and vegetables. The MOCSE13 dataset comprises four sub-datasets, each containing three types of fruits or vegetables, making it applicable for the MOC task.
Compared to the aforementioned datasets, the constructed NWPU-MOC dataset contains both RGB and NIR aerial images, along with a more extensive set of training and testing data, providing densely annotated information. The proposed NWPU-MOC dataset is better tailored to address MOC tasks in aerial scenarios.

\subsubsection{Remote sensing or aerial scenes}
Numerous datasets are constructed specifically for crowd counting in natural scenes. However, as Table \ref{tabel:1} shows, there are only a few datasets designed for object counting in remote sensing and aerial scenes.  These datasets consist of a limited number of images and are annotated with single-category objects.
For example, the Olive-Tree dataset \cite{salami2019fly} contains only 10 images, and the DLR-ACD dataset \cite{bahmanyar2019mrcnet} contains 33 images. The COWC dataset \cite{mundhenk2016large} and CARPK dataset \cite{hsieh2017drone} are annotated only for cars. Recently, the RSOC dataset \cite{gao2020counting} is constructed for remote sensing object counting, which includes four subsets: Building, Small-vehicle, Large-vehicle, and Ship, each is annotated for a single category.
Existing object counting datasets for aerial remote sensing scenes only annotate a single object category. Considering that aerial images often contain objects with multiple categories. Therefore, we collect and annotate a multi-category object counting dataset to address this gap.

\begin{figure*}[!ht]
	\centering
	\includegraphics[scale=0.65]{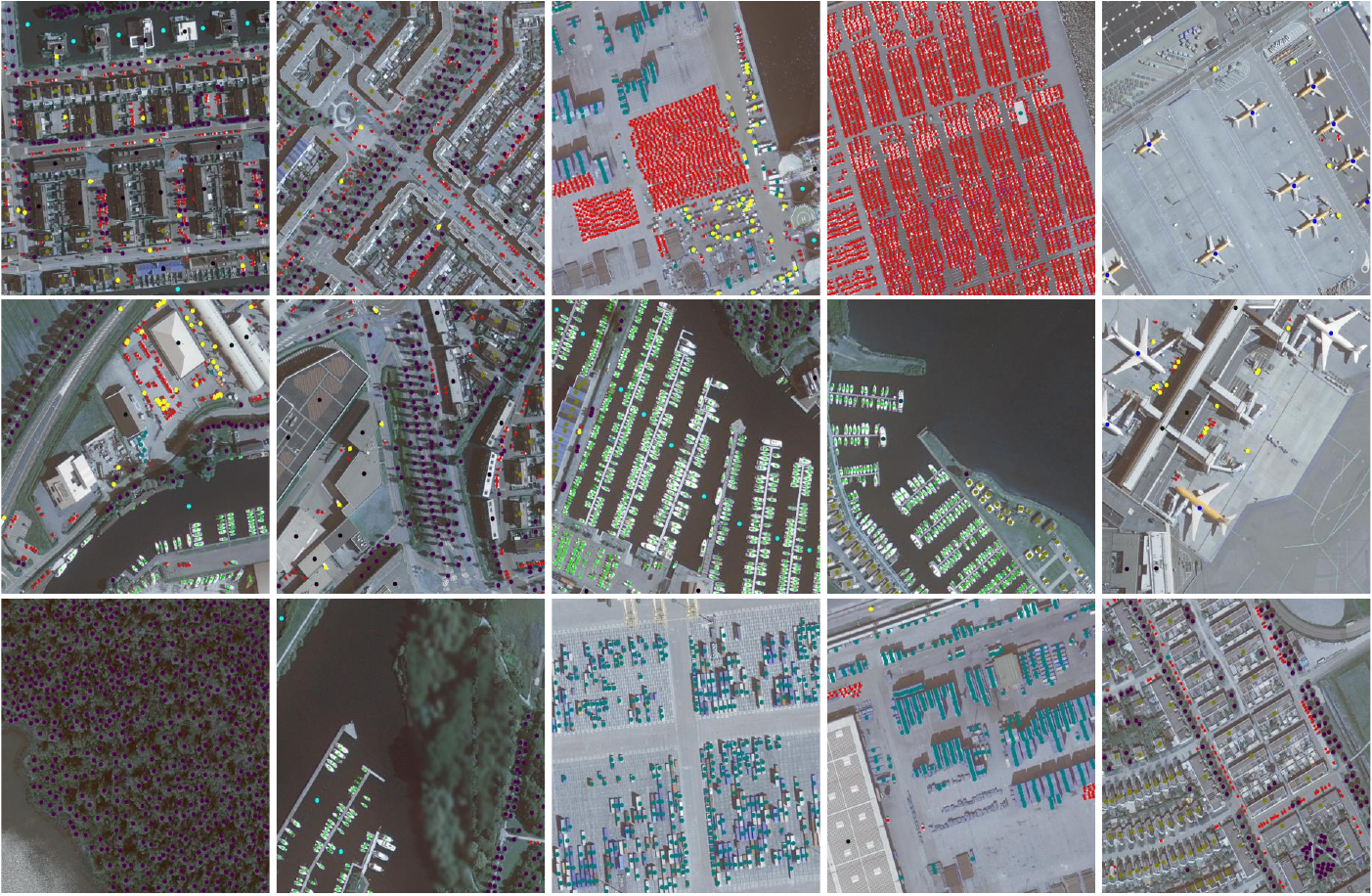}
	\caption{Examples of annotations in the NWPU-MOC dataset. Different categories of objects are labeled with center points of different colors (The color of object categories refer to Fig. \ref{fig:1}). As shown in the bottom left corner of the figure, we blurred the tree for the difficult-to-recognize by humans in the annotation, and therefore its count will be ignored in the counting algorithm.}
	\label{fig:5}
\end{figure*}

\begin{figure}[!ht] 
	\centering
	\includegraphics[scale=0.54]{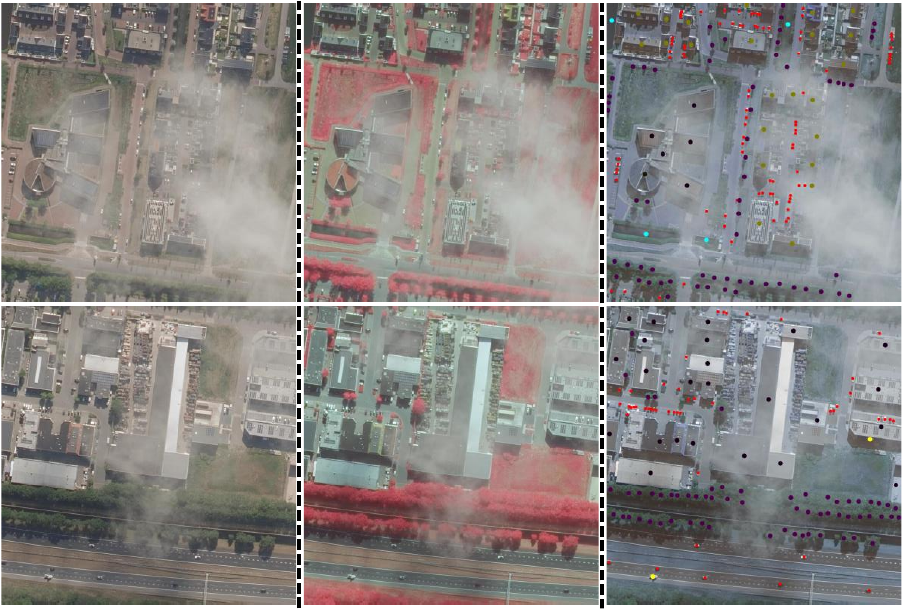}
	\caption{The first column is the RBG images, the second column is the pseudo-color images of the NIR band, and the third column is the images after annotation.}
	\label{fig:6}
\end{figure}

\subsection{Object Counting Methods}
\subsubsection{Object Counting in Natural Scenes}
Current methods for object counting can be divided into the following three categories: detection-based methods \cite{detection1,detection2,dijkstra2018centroidnet, idrees2013multi, chen2012feature}, regression-based methods \cite{Jiang_2019_CVPR, tan2011semi, liu2019context} and density map-based methods \cite{lempitsky2010learning, zhang2016single, li2018csrnet, ma2019bayesian, liu2018leveraging}. Early object counting methods are mainly detection-based methods \cite{moranduzzo2013automatic}, \cite{kamenetsky2015aerial}, which firstly detect object instances of interest and then count the number of bounding boxes. Thanks to powerful detectors, these methods achieved satisfactory performance in sparse scenes. However, in highly congested scenes, these detection-based methods are prone to failure, because the object instances are usually small in size and easily interfered with by the background.
Regression-based methods consider counting as a task of predicting global density, aiming to estimate the object count directly from the input image and the associated features. For instance, Tan \textit{et al.} \cite{tan2011semi} propose a semi-supervised elastic network.  Chan \textit{et al.} \cite{chan2009bayesian} propose a crowd-counting regression method by combining approximate Bayesian Poisson regression with a Gaussian process.
Nonetheless, it is worth noting that these regression-based approaches primarily emphasize the global features of the image, which can result in larger errors within the model and a lower tolerance for error.

The density map-based method is proposed by Lempitsky and Zisserman \cite{lempitsky2010learning}, which involves regressing a density map to estimate the number of objects. 
In recent years, many methods based on density maps designed for crowd counting tasks in dense scenes are proposed and their counting performance is significantly improved. 
MCNN is proposed by Zhang \textit{et al.} \cite{zhang2016single}, which can be fed with an image of arbitrary resolution and learn multi-scale features. Li \textit{et al.} \cite{li2018csrnet} propose a CSRNet by performing the dilated convolution, it can understand highly crowded scenes. Bayesian loss is introduced by Ma \textit{et al.} \cite{ma2019bayesian}, which abandons the density map-based method and directly utilizes point annotations for addressing this task. Density map-based methods demonstrate outstanding performance in the crowd counting task and become the mainstream methods. Recently, Liu \mbox{\textit{et al.}} \mbox{\cite{DSACA}} extended the output of the counting model to multiple layers for simultaneous estimation of vehicles and pedestrians in a single image. They also introduced a category attention module to multiply with the density map, reducing inter-class interference.

\subsubsection{Object Counting From the Remote Sensing and Aerial Viewpoint}
Aerial or remote sensing images are captured from greater distances and vertical perspectives, resulting in a wider coverage area that includes complex scene content, which raise great challenges for the existing counting models.
Bahmanyar \textit{et al.} \cite{bahmanyar2019mrcnet} introduce a crowd counting dataset based on UAV views and propose a multi-resolution network to estimate the number of pedestrians in aerial images. LPN \cite{hsieh2017drone} is proposed by using the regular spatial layout of vehicles for both vehicle counting and localization.
The tolerance repulsion principle proposed by Stahl \textit{et al.} \cite{stahl2018divide}, is an approach designed for predicting image-level counting by dividing images into a set of divisions.
Inspired by the object detectors, Li \textit{et al.} \cite{li2019simultaneously} propose using a unified framework to simultaneously detect and count vehicles. The Spatio-Temporal Neighbourhood Awareness Network  \cite{wen2021detection} further unifies the density map estimation, localization and tracking tasks in a single network. Gao \textit{et al.} \cite{gao2020counting} construct an RSOC dataset and propose using attention mechanism and deformation convolution to achieve better counting performance. Ding \textit{et al.} \cite{ding2022object} propose an adaptive density map generation algorithm to address scale variations and other challenges in remote sensing scenes.  As mentioned above, existing object counting algorithms typically focus on counting objects of a single category, which cannot be directly applied to the MOC task.

\begin{figure*}[!ht]
	\centering
	\includegraphics[scale=0.54]{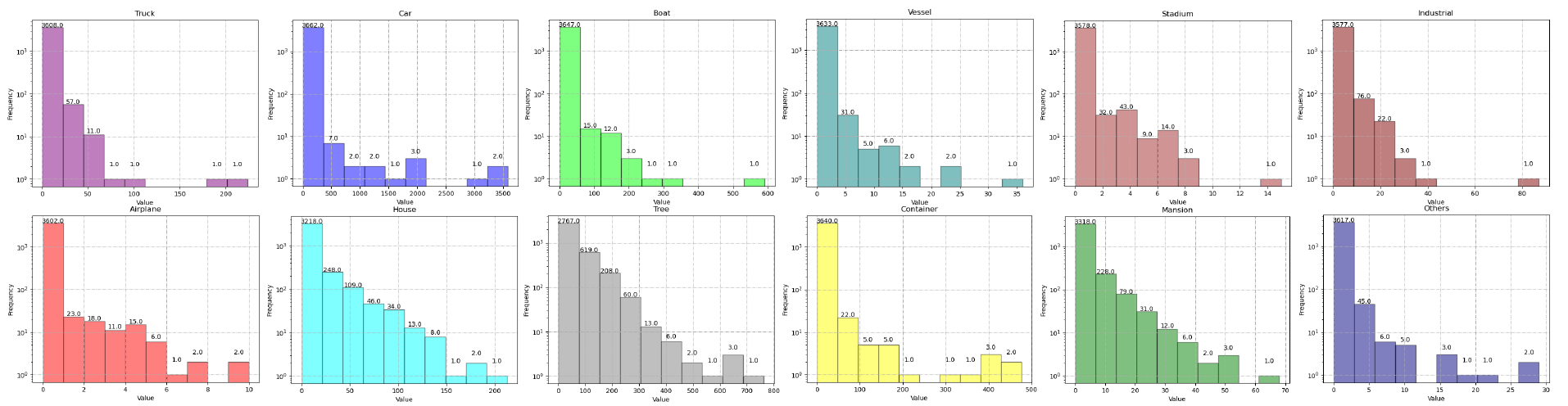}
	\caption{Histogram of the distribution of the labeled points for each category of objects in the NWPU-MOC dataset.}
	\label{fig:2}
\end{figure*}

\section{Details of the NWPU-MOC Dataset}
In this paper, our goal is to count objects with multi-category in an image simultaneously. As reviewed in Section II, the existing object-counting datasets do not apply to the MOC task. Therefore, we collect an aerial image dataset (\textbf{NWPU-MOC}), finely annotate the 14 categories of objects commonly appearing in aerial scenes. In the following, we present a detailed description of the proposed dataset, including data collection, annotation, and characterization.

\begin{figure}[!h]
	\centering
	\includegraphics[scale=0.35]{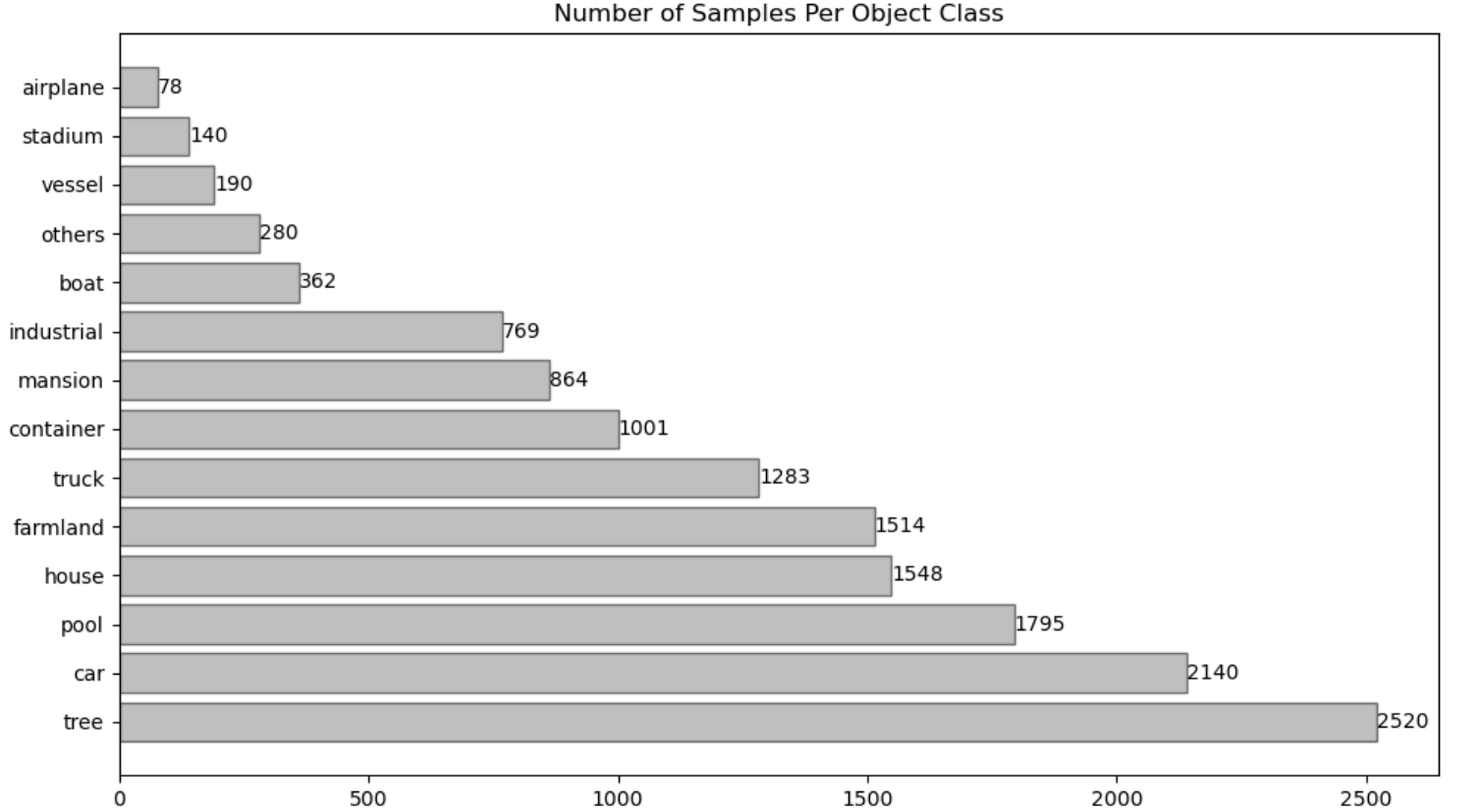}
	\caption{The statistics about the number of images for each category in the NWPU-MOC.}
	\label{fig:3}
\end{figure}
\subsection{Data Collection and Specification} 
Specifically, all aerial images are downloaded from an open-source website\footnote{https://opendata.beeldmateriaal.nl}, which provides all ortho-rectified aerial images captured by aircraft with a spatial resolution of 0.25 m/pixel over the entire Netherlands. All images contain four spectral channels: Red (R), Green (G), Blue (B), and Near-infrared (NIR). We carefully select some representative sparse scenes and dense scenes (such as ports, parking lots, blocks, industrial parks, airports, beaches, farmland, and lakes), and crop them into PNG images with a resolution of 1024 $\times$ 1024 pixels. In the end, we reserve 3,416 aerial images to make up our NWPU-MOC dataset. It is worth noting that we placed the NIR band in the first band and saved it as a pseudo-color PNG image (Fig. \ref{fig:6}).

\subsection{Data Annotation}
After analyzing and discussing the content of images in the dataset, the objects with 14 categories are selected and labeled in the images, which include Airplanes, Boats, Car, Container, Farmland, House, Industrial, Mansion, Pool, Stadium, Tree, Truck, Vessel and Others. Before annotating the images, different object categories are assigned to the 11 annotators. After the first round of fine-grained annotation, the annotation members cross-check each other's annotation categories and refine the annotation again. In the labeling process, we mark some objects that are not pre-defined as other categories. For trees occurring in clusters, the number of which may not be identified, we annotate them using boxes and process them later using blurring kernels (see Fig. \ref{fig:5}) to make the object counting algorithm ignore this part of the number. The center points of the different categories of objects appearing in an image are annotated and the coordinate ($x_{i}$, $y_{i}$) information of the annotated points is saved in a JSON format. To facilitate subsequent researchers, we use the original annotated JSON files to generate corresponding Numpy files, each containing 14 arrays with a size of 1024 $\times$ 1024, where the position of the corresponding annotated point in each array is 1, and vice versa is 0. Finally, we provide the original annotated information JSON files, the Numpy files of MOC-14 and MOC-6.
\vspace{-0.2cm}
\subsection{Data Characteristics}
In Table \ref{tabel:1}, we compare the NWPU-MOC with existing object counting datasets in aerial and remote sensing scenes. As shown in the table, NWPU-MOC provides 3416 aerial scenes. The major difference with these datasets is that NWPU-MOC simultaneously labeled 14 categories of objects with center points in an image, where the total number of labeled points is 38,3195. In addition, each scene in the NWPU-MOC dataset contains both RGB and NIR images. In aerial scenes, there may be occlusion of ground objects due to factors such as weather and lighting conditions (as shown in Fig. \ref{fig:6}). The NIR band has a longer wavelength which can provide richer discriminative features to help alleviate the occlusion problem.
Fig. \ref{fig:2} analyzes the distribution of the number of object categories of each image in the NWPU-MOC by using histograms. Fig. \ref{fig:3} shows the number of samples contained in each type of object in the NWPU-MOC. Since there is a sample imbalance among various ground objects in the real world, it can be seen from the figures that there is a long-tail distribution among sample categories in the MOC dataset. Further, in practical applications, we roughly divide the 14 categories of NWPU-MOC into 6 categories (MOC-6). As in shown Fig. \mbox{\ref{fig:4}}, Farmlands and pools actually appear as a background in the aerial images, so their number will not be counted as negative samples in MOC-6.

\subsection{Challenge}
Owing to the unique properties of aerial imagery and the particularity of multi-category object counting task, the NWPU-MOC dataset presents some challenges in the following aspects.
\subsubsection{Large-scale variation} As shown in Fig. \ref{fig:5}, due to the unique capture angles and heights of aerial images, of aerial images, there are larger-scale objects (\textit{e.g.}, airplanes, trucks, buildings) and smaller-scale objects (\textit{e.g.}, cars) in the same image. These scale differences exist not only in the same image but also between different images.
\subsubsection{Complex background} The presence of complex background information in the aerial images tends to disturb the model. And the instances of objects present in the images are subject to occlusion by vegetation, light, weather, and other factors  (as shown in Fig. \ref{fig:6}).
\subsubsection{Long-tailed distribution of Samples} The NWPU-MOC dataset exhibits a substantial distribution imbalance among categories. For instance, cars appear in 2,140 images, while only 78 images contain airplanes (as shown in Fig. \ref{fig:3}). This imbalance poses a challenge in training unbiased models using unbalanced datasets.
\subsubsection{Presence of dense scenes} Aerial images cover wide areas, with object instances occupying only a small portion of the image. In some scenes such as ports and parking lots, there are congested objects with multiple categories.  For instance, in NWPU-MOC there is an image with 3, 582 point-labeled cars. Counting multi-category dense objects within a single image presents a considerable challenge.
%\subsection{Data Split and Evaluation Protocol}

\begin{figure*}[!ht]
	\centering
	\includegraphics[scale=0.52]{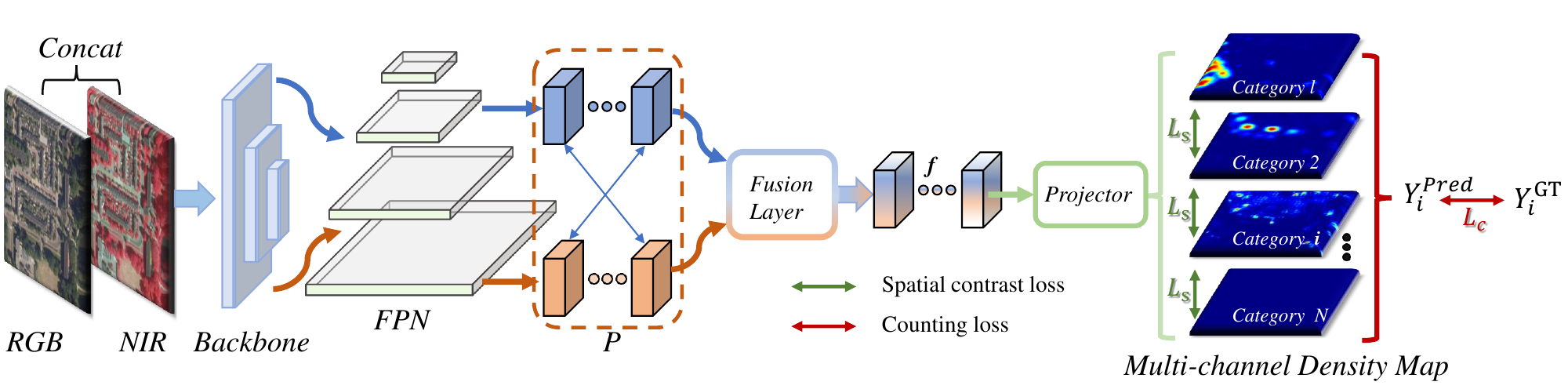}
	\caption{Overview of the proposed multi-channel density map counting framework (MCC). The arrows in the figure represent the flow of data, while the arrows in feature map \mbox{$P$} denote the NIR-RGB features after multi-scale feature fusion through FPN.}
	\label{fig:7}
\end{figure*}
\section{Methodology}
In this section, we propose a \textbf{M}ulti-\textbf{C}hannel density map \textbf{C}ounting (\textbf{MCC}) framework for multi-category object counting task. As shown in Fig. \ref{fig:7}, we use a density map-based approach in MCC.
At first, $RGB$ and $NIR$ images are taken as input and their multi-scale features are extracted by using a backbone network with pre-trained weights. Then the multi-scale feature information is fused using a feature pyramid network (FPN). Afterward, the NIR and RGB features ($f_{rgb}$ and $f_{nir}$) are fused and projected directly into a multi-channel density map ($Y^{Pred}$). During the model training phase, the \mbox{$Y^{Pred}$} is drawn close to the generated density map (\mbox{$Y^{GT}$}) by optimizing the counting loss \mbox{$\mathcal{L}_{\mathcal{C}}$}. Meanwhile, we propose a spatial contrast loss \mbox{$\mathcal{L}_{\mathcal{S}}$} as a penalty for the overlap between the same spatial locations of each channel in \mbox{$Y^{Pred}$}. The \mbox{$\mathcal{L}_{\mathcal{S}}$} is optimized to model the mapping relationships between channels in \mbox{$Y^{Pred}$} and each category.

\subsection{Overall Structure of MCC}
The proposed MCC is divided into two stages: in the density map generation phase, a Gaussian density map $Y^{GT}$ with $N\mbox{-}channel$ is generated using the corresponding  dots labeled maps $D$ (containing $N$ channels of dots labels for each object category) of the input image $I$ by convolution with a Gaussian kernel,
\begin{equation}
    Y_{i}^{GT}=D_{i}(x) * G(x, \sigma),  \qquad  i=1,2,...,N , 
   \label{eq:1}
\end{equation}
\begin{equation}
	   G(x, \sigma)=\frac{1}{2 \pi \sigma^{2}} e^{-\frac{\left\|x-x^{\prime}\right\|^{2}}{2 \sigma^{2}}},
\end{equation}
where $G(x, \sigma)$ denotes the 2D convolution kernel with variance $\sigma$ and kernel size $k$. 
After that, the generated Gaussian density map \textbf{$Y^{GT}$ } is used as supervised information in the density map prediction stage. Then, a backbone network (here we use the swin transformer \cite{swin}) is used to extract multi-scale feature information from $I_{rgb}$ and $I_{nir}$,
\begin{equation}
		Y = Concat(I_{rgb}, I_{nir}), 
\end{equation}
\begin{equation}
		F_1,F_2,F_3 = \operatorname{ST}(Y),
\end{equation}
where $Concat(\cdot)$ denotes the operation of concatenation,  $Y  \in \mathbb{R} ^ {4  \times H \times W}$, $F_1  \in \mathbb{R} ^ {C \times H/4 \times W/4}$, $F_2 \in \mathbb{R} ^ {2C \times H/8 \times W/8}$ and $F_3 \in \mathbb{R} ^ {4C \times H/16 \times W/16}$ ($C$=128), respectively. To alleviate the scale variation problem present in aerial scenes, following the previous work \cite{SwinCounter}, we fuse multi-scale feature information by using a feature pyramid network(FPN),
\begin{equation}
	\begin{array}{l}
		P_1,P_2,P_3 = \operatorname{FPN}(F_1,F_2,F_3).
	\end{array}
\end{equation}
After this, $P_1  \in \mathbb{R} ^ {C \times H/4 \times W/4 } $, $P_2 \in \mathbb{R} ^ {C \times H/8 \times W/8}$ and $P_3 \in \mathbb{R} ^ {C \times H/16 \times W/16}$  are concatenated in the feature dimension using an up-sampling operation to scale them to the same scale.
\begin{equation}
	\begin{array}{l}
		P = Concat(P_1,\mathcal{F}_{U P\times2}(P_2),\mathcal{F}_{U P\times4}(P_3)),
	\end{array}
\end{equation}
where $\mathcal{F}_{U P\times2}(\cdot)$ and $\mathcal{F}_{U P\times4}(\cdot)$ denote 2$\times$ up-sampling and 4$\times$ up-sampling, respectively.  \mbox{$P \in \mathbb{R} ^ {3C \times H/4 \times W/4}$} contains the RGB and NIR feature information after feature pyramid fusion. As shown in  Fig. \mbox{\ref{fig:7}}, the proposed method fuses the features of both RGB and NIR spectra, using the complementary properties of the two spectra to alleviate problems such as background darkness and occlusion that may exist in RGB aerial images,
\begin{equation}
	\begin{array}{l}
		f =\mathcal{F}_{RGB\&NIR}(P).
	\end{array}
\end{equation}
The feature map $f$ after NIR and RGB fusion is projected directly (here we use a $1\times1$ convolutional layer) to the $N$ channels density map $Y^{Pred}$,
\begin{equation}
	\begin{array}{l}
		Y^{Pred} =Projector(f),
	\end{array}
\end{equation}
where $Y^{Pred} \in \mathbb{R} ^ { w \times h \times N}$. Finally, counting loss $\mathcal{L}_{\mathcal{C}}$ is calculated using the generated Gaussian density map as supervised information with the predicted density map,
\begin{equation}
	\begin{array}{l}
	\mathcal{L}_{\mathcal{C}}=\left\|Y_{i}^{GT} - Y_{i} ^{Pred}\right\|_{2}^{2}.
	\end{array}
\end{equation}

Following the previous works\cite{gao2020counting, SwinCounter}, the MSE loss function is used as the counting loss. However, for the MOC task, different from single-category object counting, there could be overlaps in the same spatial locations of each channel of predicted density maps. In this paper, we propose a spatial contrast loss function $\mathcal{L}_{\mathcal{S}}$ to Minimize the similarity of the spatial location between predicted density maps.
The final loss function is as follows,
\begin{equation}
	\begin{array}{l}
		\mathcal{L} =\mathcal{L}_{\mathcal{C}} + \gamma  \mathcal{L}_{\mathcal{S}},
	\end{array}
\end{equation}
where $\gamma$ is used as a hyper-parameter to tune the weights of $ \mathcal{L}_{\mathcal{S}}$ for its joint optimization with $\mathcal{L}_{\mathcal{C}}$.
\begin{figure}[!h]
	%是可选项 h表示的是here在这里插入，t表示的是在页面的顶部插入
	\centering
	\includegraphics[scale=0.63]{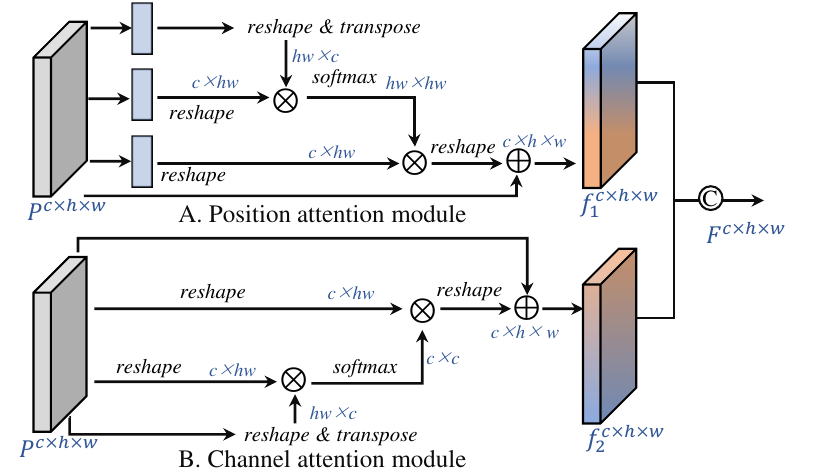}
	\caption{Dual attention for RGB-NIR fusion.}
	\label{fig:8}
\end{figure}
\subsection{RGB and NIR Fusion Layer}
As shown in Fig. \ref{fig:8}, the RGB-NIR features $P \in \mathbb{R} ^ {  c \times h \times w }$ are fed into a position attention module \cite{fu2019dual}.
Firstly, $P$ is transformed into 3 feature maps ($P_1, P_2, P_3$) with different shapes by reshape after using three $1\times1$ convolutional layer,

\begin{equation}
	\begin{array}{l}
		P_{1}^{c \times hw} =\mathcal{R}\left(\mathcal{K}_{1 \times 1}(P)\right), \\
		P_{2}^{c \times hw} = \mathcal{R}\left(\mathcal{K}_{1 \times 1}(P)\right), \\
		P_{3}^{c \times hw} = \mathcal{R}\left(\mathcal{K}_{1 \times 1}(P)\right),
	\end{array}
\end{equation}
where $\mathcal{K}_{1 \times 1}$ denotes a convolutional layer with kernel size of $1 \times 1$ and $\mathcal{R}$ represents the reshape operation. After that, we use the transpose matrix of $P_1$ and $P_2$ for matrix multiplication, and use a softmax layer to calculate the spatial attention map  $S \in \mathbb{R} ^ {hw \times hw}$,
\begin{equation}
	\begin{array}{l}
	S^{j i}=\frac{\exp \left(P_{1}^{i} \cdot P_{2}^{j}\right)}{\sum_{i=1}^{hw} \exp \left(P_{1}^{i} \cdot P_{2}^{j}\right)} .
	\end{array}
\end{equation}
Later, the transpose of $S$ and $P_3$ is matrix multiplied and the result is reshaped as  $\mathbb{R} ^ {  c \times h \times w }$, and finally summed with a matrix $p$. The process can be formulated as,
\begin{equation}
	\begin{array}{l}
		f_{1}^{j}=\alpha \sum_{i=1}^{hw}\left(S^{j i} \cdot P_{3}^{i}\right)+ P^{j} ,
	\end{array}
\end{equation}
where $\alpha$ is initialized as 0 and is gradually learned during the training process. As shown in Fig. \ref{fig:8}.B, we use a channel attention mechanism \cite{fu2019dual} to fuse RGB-NIR features in the channel dimension. Different from the position attention module, here we do not use the convolution operation for $P$, but directly reshape $P$, and then compute the attention map for itself,
\begin{equation}
	\begin{array}{l}
		P_{4}^{c \times hw} = \mathcal{R}\left(P\right).
	\end{array}
\end{equation}
We use the transpose of matrix $p$ and $p$ to do matrix multiplication and use a softmax to calculate the channel attention map $C \in \mathbb{R} ^ {c \times c}$,
\begin{equation}
	\begin{array}{l}
		C^{j i}=\frac{\exp \left(P_{4}^{i} \cdot P_{4}^{j}\right)}{\sum_{i=1}^{hw} \exp \left(P_{4}^{i} \cdot P_{4}^{j}\right)} ,
	\end{array}
\end{equation}
where $C^{j i}$ indicates the correlation of $C^{j i}$ and $P_{4}^{j}$ in the channel dimension. Then, $C$ is used to multiply with the transpose of $P_{4}^{i}$ and the calculation result is reshaped to the same shape as $P_{4}^{i}$, and finally summed with $P_{4}^{i}$ to get $f_2$,
\begin{equation}
	\begin{array}{l}
		f_{2}^{j}=\beta \sum_{i=1}^{\textsc{C}}\left(C^{j i} \cdot P_{4}^{i}\right)+ P^{j}.
	\end{array}
\end{equation} 
Finally, $f_1$ and $f_2$ are concatenated and their feature dimensions are reduced to $c$ by using a $1\times1$ convolution layer to obtain $F \in \mathbb{R} ^ {c \times h \times w}$,
\begin{equation}
	\begin{array}{l}
		F = \mathcal{K}_{1 \times 1}(Concat(f_{1}, f_{2})).
	\end{array}
\end{equation}
\subsection{Spatial Contrast Loss}
As mentioned above, multi-category object counting (MOC) involves the task of distinguishing between different categories of objects and accurately counting the objects within each category. This requires a one-to-one mapping relationship between the channels of the regressed density maps and the object categories.
The previous supervised paradigm optimizes the counting loss function alone, which is not sufficient to capture the one-to-one mapping relationship between the predicted density map and the object categories. To address this issue, we propose a spatial contrast loss \mbox{$\mathcal{L}_{\mathcal{S}}$}, which is jointly optimized with the counting loss \mbox{$\mathcal{L}_{\mathcal{C}}$}.

For each channel $Y_i^{Pred} \in \mathbb{R}^{w \times h \times 1}$ in the predicted density map $Y^{Pred} \in \mathbb{R}^{w \times h \times N}$, we flatten it into a 1-D vector $u_i$ of length $h \times w$,

\begin{equation}
u_i = \text{flatten}(Y_i^{Pred}), \quad i = 1,2,...,N,
\end{equation}
where $u_i$ represents the spatial vector of $Y_i^{Pred}$. Next, we define $M_{ij}$ as the cosine similarity between the spatial vectors $u_i$ and $u_j$,

\begin{equation}
M_{ij} = S_{\theta}(u_{i}, u_{j}), 
\end{equation} 

\begin{equation}
S_{\theta}(u_{i}, u_{j}) = \frac{\langle u_{i}, u_{j} \rangle}{\|u_{i}\|_{2} \cdot \|u_{j}\|_{2}},
\end{equation}
where $\langle \rangle$ denotes the dot product between $u_i$ and $u_j$, and $||.||_{2}$ denotes the $L2$-norm of the vectors. This process generates a spatial similarity matrix $M$ for each $Y_i^{Pred}$,

\begin{equation}
M = \left[
\begin{array}{cccc}
S_{\theta}(u_1,u_1) & S_{\theta}(u_1,u_2) & \cdots & S_{\theta}(u_1,u_N)\\
S_{\theta}(u_2,u_1) & S_{\theta}(u_2,u_2) & \cdots & S_{\theta}(u_2,u_N)\\
\vdots & \vdots & \ddots & \vdots \\
S_{\theta}(u_N,u_1) & S_{\theta}(u_N,u_2) & \cdots & S_{\theta}(u_N,u_N) 
\end{array}
\right].
\end{equation}

Finally, the spatial contrast loss $\mathcal{L}_{\mathcal{S}}$ is calculated as the mean value of each element in the matrix $M$,

\begin{equation}
\mathcal{L}_{\mathcal{S}} = \overline{M}_{ij}, \quad i,j=1,2,...,N.
\end{equation}

\begin{table*}[!ht]
	\centering
	\caption{Comparison of our method with other advanced counting methods on MOC-6 datasets. The best and second best results are marked in the table by \textbf{bold} and \underline{underlined}, respectively.}
	\resizebox{\textwidth}{!}{
		\begin{threeparttable}
			\begin{tabular}{l |c c c c c c c c c c c c c | c c }
				\Xhline{1.2pt}
    
				\multirow{2}{*}{Methods} &  \multirow{2}{*}{ ($\sigma$,$s$) } & \multicolumn{2}{c}{Ship}&\multicolumn{2}{c}{Vehicle} & \multicolumn{2}{c}{Building} & \multicolumn{2}{c}{Container} & \multicolumn{2}{c}{Tree} & \multicolumn{2}{c|}{Airplane} &\multicolumn{1}{c}{\multirow{2}{*}{$\mathrm{\overline{MSE}}$}} &\multirow{2}{*}{WMSE} \\ 
				~ & ~ & MAE & RMSE & MSE  & RMSE  & MAE & RMSE & MAE  &RMSE  & MAE & RMSE & MAE  & RMSE &  &   \\ \hline
				
				\multirow{2}{*}{MCNN \cite{zhang2016single}} & (2,5) & 3.5121 & 16.5731 & 14.0003 & 28.8503 & 8.8326 & 16.6600 & 5.0883 & 24.8989 & 26.4277 & 45.2319 & 0.0615 & 0.4768 & 22.1152 & 125.8497  \\ 
				 ~ & (4,15) & 2.4634  & 17.7363 & 18.6045 & 44.1941 & 8.1840 & 16.8480 & 4.8467 & 25.1023 & 28.8071 & 46.3297 & 0.0615 & 0.4768 & 25.1145 & 171.2414  \\  
				 \hline
				\multirow{2}{*}{CSRNet \cite{li2018csrnet}} & (2,5) & 3.8120 & 13.5363 & 6.0453  & 13.5363 & 7.0123 & 13.6163 & 5.3514 & 25.4511 & 17.9114 & 35.6847 & 0.094 & 0.4717 & 17.7017 & 73.0056 \\ 
				~ & (4,15) & 4.6024 & 17.4716 & 13.7151 & 24.3899 & 8.0835 & 14.8328 & 6.5580 & 25.4806 & 27.8693 & 47.7390 & 0.2623 & 0.5090 & 21.7371 & 124.1803  \\  
				\hline
				\multirow{2}{*}{SFCN \cite{SFCN}}& (2,5) & 1.0987 & 4.2537 & 5.6233 & 22.4487 & 3.3484 & 6.6719 & 4.4195 & 26.1628 & 15.3042 & 31.4514 & 0.0615 & 0.4768 & 15.2442 & 73.3208 \\
				~ & (4,15) & 1.7080 & 10.5572 & 5.5301 & 13.8747 & 4.0828 & 8.3646 & 1.8993 & 7.3178 & 15.8311 & 32.2336 & 0.0615 & 0.4768 & 14.8041 &  77.6888 \\  
				\hline
				\multirow{2}{*}{SCAR \cite{scar}} & (2,5) & 0.7119 & 3.681 & 4.7557 & 21.5266 & 3.0669 & 6.4391 & 1.8979 & 8.1291  & 15.4480 & 32.1106 & 0.0615 & 0.4768 & 12.0605 & 48.6011\\
				~ & (4,15) & 0.8991 & 4.6874 & 6.7638 & 45.0773 & 3.3591 & 6.8968 & 1.8298 & 7.1641 & 16.4259 & 34.1955 & 0.0615 & 0.4768 & 16.4163 & 120.3997  \\  
				% \hline
				% \multirow{2}{*}{CAN \cite{CAN}} & (2,5) & 1.4704 & 4.0821 & 8.2740 & 26.0846 & 3.8957 & 7.0062 & 1.8885 & 5.9988 & 17.5251 & 34.0307 & 0.1104 & 0.4708 & 12.9455 & 57.6855 \\
				% ~ & (4,15) & 1.4100 & 3.8011 & 7.1397 & 48.7376 & 3.3612 & 6.8092 & 2.4694 & 8.9163 & 15.8042 & 32.7978 & 0.0765 & 0.4750 & 16.9228 & 127.6252  \\  				
%				SFANet \cite{SFANet} & & & & & & & & & & & & & &  \\
				\hdashline
				\multirow{2}{*}{ASPNet \cite{gao2020counting}} & (2,5) & 2.6420 & 5.7211 & 6.1409 & 17.0018 & 3.9111 & 7.4463 & 2.4136 & 8.8967 & 15.9353 & 32.8638 & 0.0945 & 0.4728 & 12.0671 & 42.3413 \\
				~ & (4,15) & 2.0851 & 6.1556 & 11.3318 & 35.6601 & 3.9618 & 6.8671 & 2.5618 & 8.0817 & 18.5355 & 32.4692 & 0.1105 & 0.4736 & 14.9512 & 111.2808  \\  
				\hline
				\multirow{2}{*}{PSGCNet \cite{PSGCNet}} & (2,5) & 4.0871 & 17.5694 & 7.4187 & 17.9339 & 13.8796 & 21.2722 & 6.4821 & 25.7475 & 18.5152 & 35.3097 & 0.1142 & 0.4723 & 19.7175 & 88.3311 \\ 
				~ & (4,15) & 3.7138 & 17.5604 & 10.1730 & 18.5158 & 9.8801 & 18.9321 & 5.8081 & 25.7689 & 21.2395 & 40.2752 &  0.1204 & 0.4765 & 20.2548 & 82.2757  \\  
				\hline
				\multirow{2}{*}{SwinCounter \cite{SwinCounter}} & (2,5) & 0.8777 & 4.2339 & 4.5300 & 12.2886 & 2.7377 & 5.7930 & 2.3717 & 10.1307 & 18.3741 & 34.0095 & 0.2423 & 0.4137 & 11.1449 & 44.456 \\
                 ~ & (4,15) & 0.8212 & 3.6152 & 5.0925 & 19.1911 & 3.4919 & 7.1170 & 1.8515 & 6.3665 & 15.9580 & 33.9808 & 0.1666 & 0.4471 & 11.7863 & 45.5477  \\  
				\hline
%				PSGCNet \cite{PSGCNet}& & & & & & & & & & & & & & \\ 
                
                \multirow{2}{*}{DSACA \cite{DSACA}}& (2,5)  & 2.4642 & 17.7364 & 10.6751 & 22.9972 & 11.9554 & 23.9831 & 4.4206  & 26.1633 & 27.8939 & 54.9931 & 0.0692 & 0.4780 & 20.1546 & 63.2671 \\
                % \cellcolor{red!40}                
				                  ~ & (4,15) & 1.5117  & 4.0399 & 6.2882 & 38.7013 & 3.6883 & 7.2121 & 3.5049 & 9.9215 & 17.3888 & 32.9396 &0.1100  & 0.4757 & 15.5483 & 97.0073  \\  
				\hline
               
				\multirow{2}{*}{\textbf{MCC} (Ours)}& (2,5) & 1.0052 & 4.9582 & 4.1862 & 10.7381 & 2.9002 & 6.2173 & 1.9558 & 7.7771 & 15.2229 & 32.0804 & 0.0615 & 0.4768 & \underline{11.0374} & 42.0919 \\
                % \rowcolor{red!40}
				~ & (4,15) & 0.9684 & 4.3597 & 4.1432 & 12.7091 & 3.4963 & 7.3538 & 1.6867 & 6.8366 & 16.7922 & 35.3594 & 0.0615 & 0.4768 & 11.4826 & 43.1082  \\  
				\hline
				\multirow{2}{*}{\textbf{MCC+} (Ours) }& (2,5) & 1.0983 & 5.0885 & 4.0087 & 11.6656 & 2.8425 & 6.0414 & 1.7099 & 6.3283 & 15.3271 & 32.4639 & 0.0615 & 0.4768 & \textbf{10.3441} & \textbf{37.8798}  \\
				~ & (4,15) & 0.6935 & 3.7116 & 4.5118  & 10.9049 & 3.9028 & 8.5099 & 1.7979 & 8.2696 & 16.6845 & 35.8569 & 0.0615 & 0.4768 & 11.2883 & \underline{39.1762}  \\  
				\hline
				\Xhline{1.2pt}
			\end{tabular}
		\end{threeparttable}
	}
	\label{table:2}
\end{table*}

\begin{figure*}[!ht]
	%是可选项 h表示的是here在这里插入，t表示的是在页面的顶部插入
	\centering
	\includegraphics[scale=0.58]{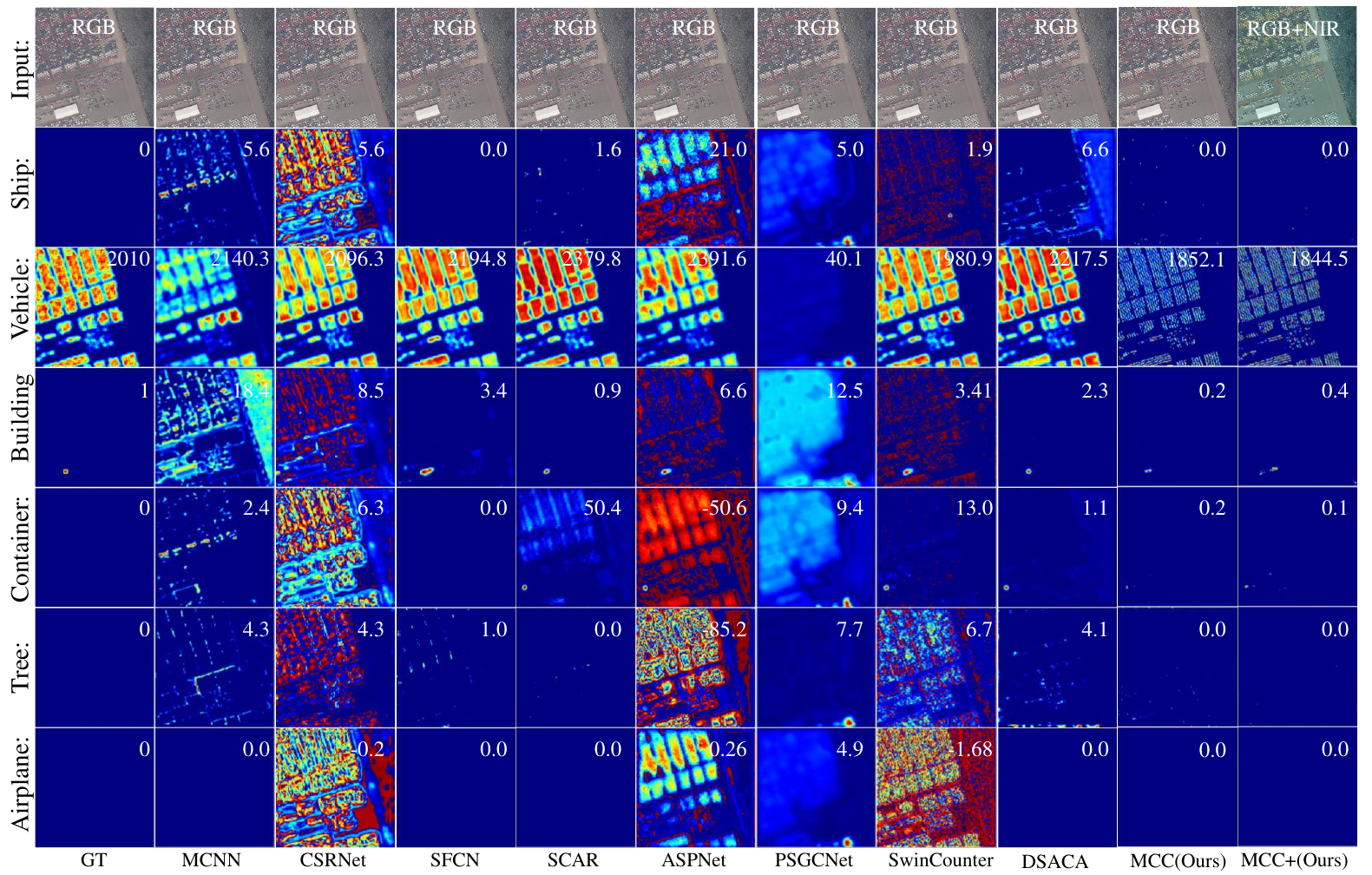}
	\caption{Visual comparison results of multi-category object counting. The first row is the original RGB image and the NIR false color image.}
	\label{fig:9}
\end{figure*}

\section{Experiments}
In this section, the NWPU-MOC dataset is used as a benchmark to evaluate the performance of our proposed method. We quantitatively compare the proposed method with other classical counting methods as well as recently designed remote sensing object counting algorithms. We also provide visual comparisons of the counting results in density maps.
To effectively evaluate the performance of different models in the MOC task under category-imbalanced distributions,  a new evaluation metric is presented. 
Furthermore, we conduct ablative experiments to analyze the effectiveness of the main components of our proposed method, including the fusion of NIR and RGB features, the spatial contrast loss function, and the key hyper-parameters.
% In this section, we evaluate the performance of our proposed method using the NWPU-MOC dataset as a benchmark. We compare our results with other classical counting algorithms as well as recently designed for remote sensing object-counting algorithms.
% To effectively evaluate the performance of different models in the MOC task under category-imbalanced distributions, a new evaluation metric is proposed. 
% Additionally, we conduct ablation experiments to analyze the effectiveness of the main components of our proposed method, including the loss function and the main hyper-parameters.
\subsection{Experimental Protocol}
\subsubsection{Data Processing}As shown in Fig. \mbox{\ref{fig:4}}, in the constructed NWPU-MOC dataset, we provide annotation information of 14 categories ground objects (MOC-14), of which we roughly classify into 6 categories (MOC-6). Due to the low spatial resolution of 0.25 m/pixel in the aerial images of NWPU-MOC, accurately counting all 14 fine-grained categories is quite challenging. In our experiments, we focused on counting objects within the roughly partitioned 6 categories (MOC-6). In the density map generation stage, we use two types of Gaussian kernels with $\sigma$=2, $s$=5 and $\sigma$=4, $s$=15 (Here, $\sigma$ and $s$ represent the bandwidth and size of the Gaussian kernel in Equation \ref{eq:1}, respectively.) to generate the Gaussian density maps by convolving with the labeled point maps. RGB and NIR images are provided with a PNG file format, where the NIR is saved in the first channel of the PNG image. Before training, the RGB and NIR images in the dataset are loaded and scaled to a uniform size of 512 $\times$ 512. We use various data enhancement operations for RGB and NIR images, including random cropping, random flipping, and normalization to enhance the generalization of the model.
\subsubsection{Experiment Setup}
Our code is built on Pytorch and $C^3$ framework \cite{C3}. All experiments are implemented on Ubuntu 18.04 OS with a single NVIDIA GTX3090 GPU. All the methods compared in our experiments are based on publicly available code and original paper to reproduce. Since the above-compared methods are used for single-category object counting tasks, in order to be applicable to multi-category object counting, we direct to output a multi-channel density map at the last convolutional layer of the models. In the training phase of the model, the Adamw optimizer is used, and set $\beta_{1}=0.9$, $\beta_{2}=0.999$, and $weight decay=10^{-2}$. The learning rate at the beginning of the training is initialized to $10^{-5}$, and then each epoch declines with a weight of $ 0.995$. All compared models were trained on the NWPU-MOC dataset with 200 epochs using the same training strategy and hyper-parameters.
\subsubsection{Evaluation Metrics}
Following previous works, Mean Absolute Error (MAE) and Root Mean Squared Error (RMSE) are commonly used for single-category object counting to evaluate the performance of the model. We report the MAE and RMSE for counting errors of each category. They are defined as follows,
\begin{equation}
	\begin{array}{l}
		\mathrm{MAE}=\frac{1}{n} \sum_{i=1}^{n}\left|\hat{X}_{i}-X_{i}\right|, 
	\end{array}
\end{equation}
\begin{equation}
	\begin{array}{l}
		\mathrm{RMSE} = \sqrt{\frac{1}{n} \sum_{i=1}^{n}\left|\hat{X}_{i}-X_{i}\right|^{2}},
	\end{array}
\end{equation}
where $n$ denotes the number of images, $X_{i}$ represents the ground truth count of objects in the $i$-th image, and $\hat{X}{i}$ represents the predicted count (sum of densities in $Y{i} ^{Pred}$). respectively. To assess the counting performance across multiple categories, we introduce the inter-category Average MSE ($\mathrm{\overline{MSE}}$),
\begin{equation}
\mathrm{\overline{MSE}} = \frac{1}{N} \sum_{i=1}^{N} \mathrm{MSE_{i}},
\end{equation}
where $N$ denotes the number of object categories, and $\mathrm{MSE_{i}}$ represents the mean squared error for category $i$.

The NWPU-MOC dataset exhibits category imbalance, meaning that each category has a different proportion of samples. Therefore, using equal weights for each category in the $\mathrm{\overline{MSE}}$ metric may not capture the bias of counting models towards specific object categories. Motivated by SoftmaxWithLoss \cite{Eigen_2015_ICCV}, we propose a Weighted MSE (WMSE) to provide an unbiased evaluation of multi-category object counting models. The WMSE assigns different weights to the MSE based on the proportion of object categories,
\begin{equation}
\mathrm{WMSE} = \frac{1}{N} \sum_{i=1}^{N} (w_i \times \mathrm{MSE_{i}}),
\end{equation}
where $w_i$ and $\mathrm{MSE_{i}}$ represent the weight and MSE error for category $i$, respectively. The weight $w_i$ for each category is calculated using the formula,
\begin{equation}
    \begin{array}{l}
	   1 / fr_i =  \ln{\frac{C_i}{M(C) }} ,\\   \\
  
        w_i = \frac{\exp \left( fr_i \right)}{\sum_{i=1}^{N} \exp \left( fr_i \right)},
    \end{array}
\end{equation}
where $M(C)$ represents the median count of objects across all categories in an image, and $C_i$ represents the count of objects for category $i$. Subsequently, the frequency $fr_i$ for each category is normalized using softmax to obtain the corresponding weight $w_i$. By assigning greater weights to less frequent object categories, the proposed WMSE can better evaluate the generalization capabilities of different counting models for long-tailed distribution issues in the context of MOC tasks.

\begin{figure*}[!ht]
	%是可选项 h表示的是here在这里插入，t表示的是在页面的顶部插入
	\centering
	\includegraphics[scale=0.58]{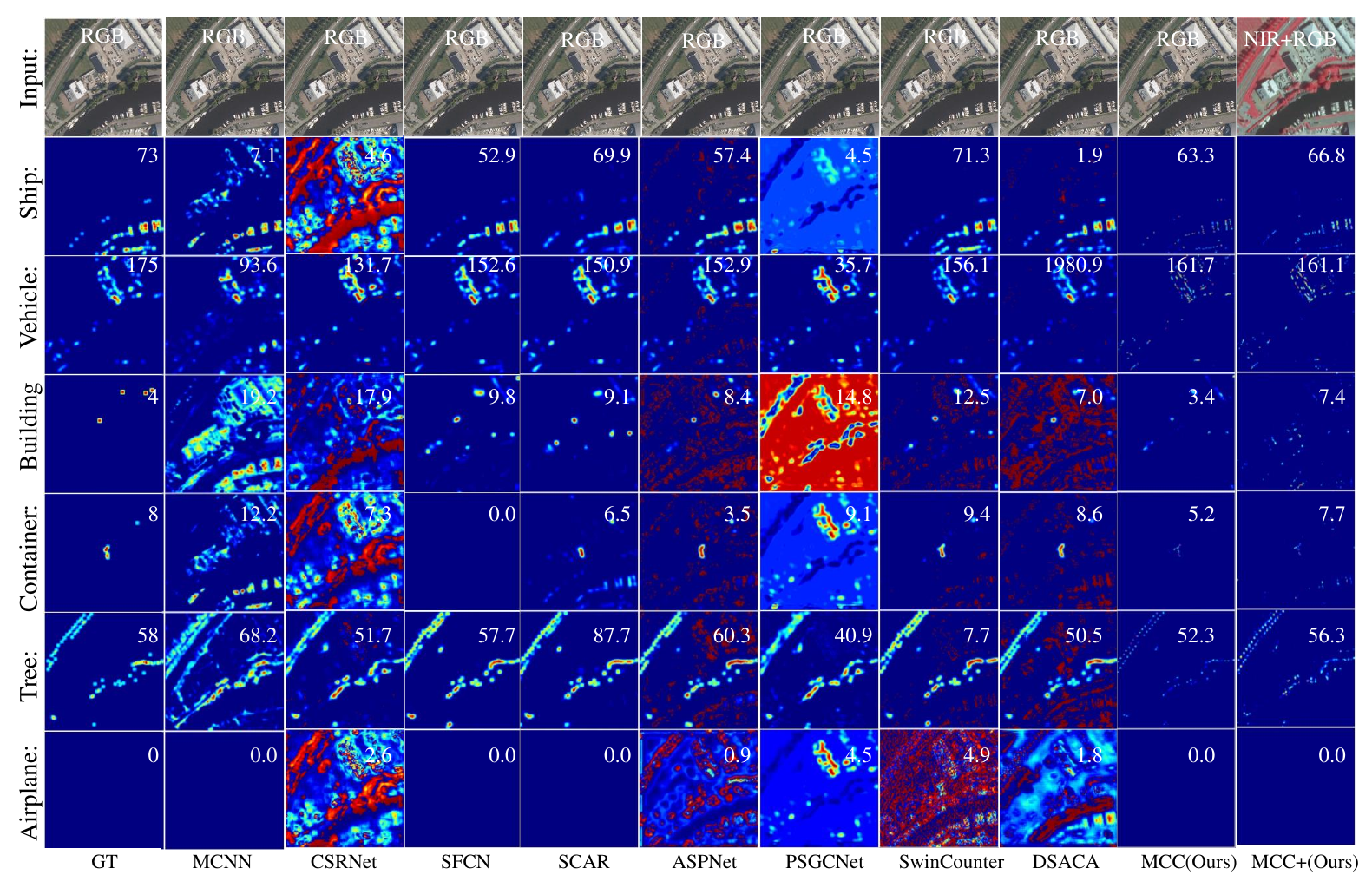}
	\caption{Visual comparison results of multi-category object counting. The first row is the original RGB image and the NIR false color image.}
	\label{fig:10}
\end{figure*}

\subsection{Mainstream Methods Involved in Evaluation}
To provide comprehensive benchmarks, we evaluate some popular object-counting methods on the NWPU-MOC dataset. These methods include the crowd-counting method (MCNN \cite{zhang2016single}, CSRNet \cite{li2018csrnet}, SFCN \cite{SFCN}, SCAR \cite{scar}) and the object-counting methods designed for remote sensing scenes (ASPNet \cite{gao2020counting}, PSGCNet \cite{PSGCNet}, SwinCounter \cite{SwinCounter}), respectively. The experimental results on the NWPU-MOC dataset, where two types of Gaussian kernels are used to generate Gaussian density maps, are presented in Table \ref{table:2}. For a fair comparison with other methods, we show the testing performance of the proposed MCC when using RGB alone, and when using RGB-NIR (where we mark the model with a symbol ``+" ). From the table, it is evident that when counting six different object categories, the proposed MCC outperforms the compared methods significantly in terms of MSE and RMSE, when using only RGB as input. Furthermore, when using both NIR and RGB as inputs, evaluation metrics such as MAE and RMSE show a significant improvement. This validates that NIR can provide more discriminative feature information, aiding in enhancing the counting performance of the model. Moreover, Table \ref{table:2} displays the MAE and RMSE errors for the categories with a smaller proportion in the dataset, which are relatively low. Consequently, using these metrics for evaluation makes it challenging to compare the counting performance between algorithms.
To address this issue, we report $\mathrm{\overline{MSE}}$ and WMSE as evaluation metrics for all categories in the MOC-6 dataset. As shown in the table, our method achieves lower $\mathrm{\overline{MSE}}$ and WMSE errors compared to other methods. By employing a training strategy that generates ground truth using two types of Gaussian kernels, our method achieves the lowest MSE counting error among the compared methods when $\sigma$ = 2 and $s$ = 5. Additionally, all the compared methods also achieve better counting performance when using smaller bandwidth ($\sigma$) and size ($s$) of the Gaussian kernels.

\begin{figure*}[!ht]
	%是可选项 h表示的是here在这里插入，t表示的是在页面的顶部插入
	\centering
	\includegraphics[scale=0.58]{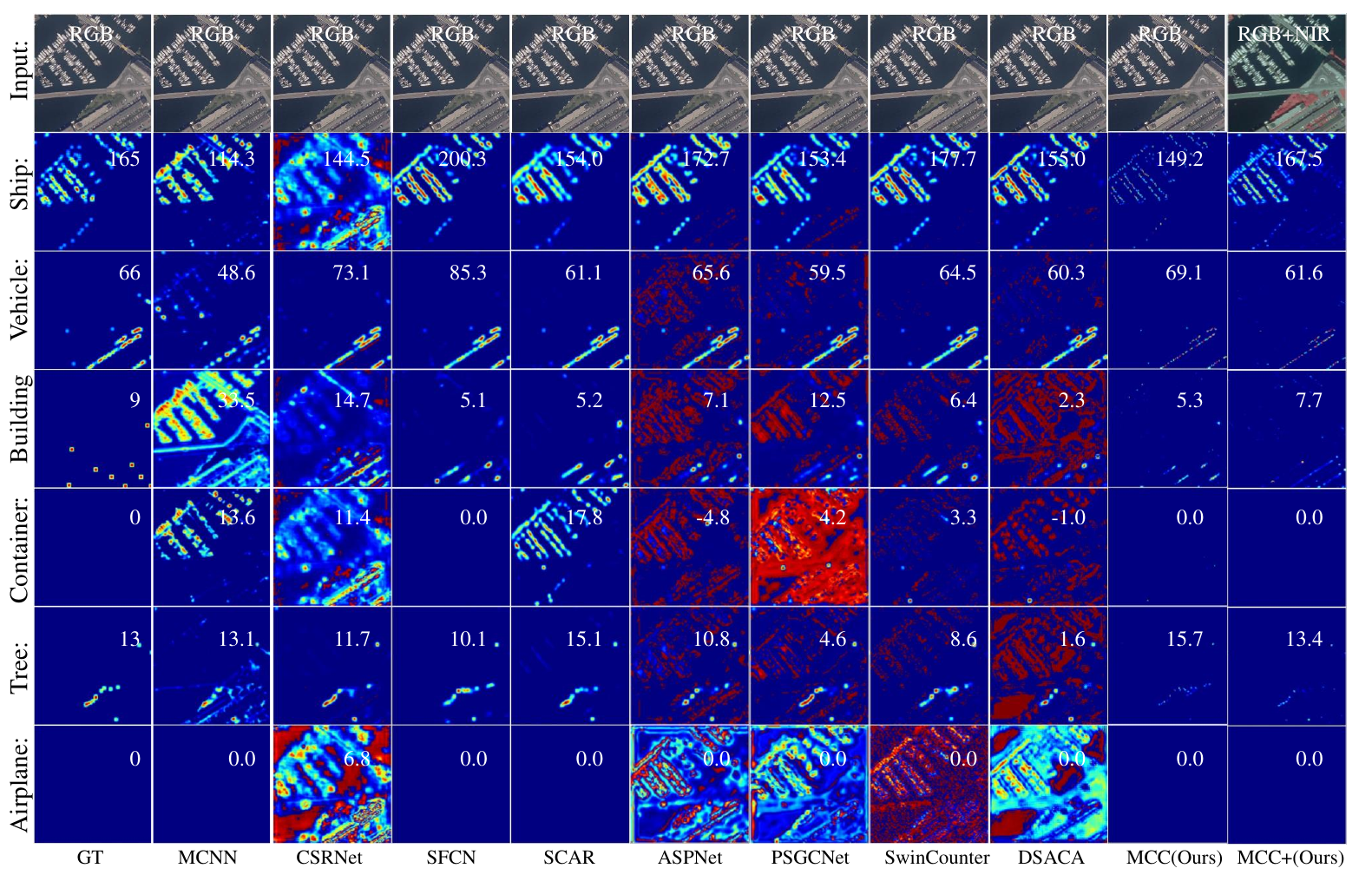}
	\caption{Visual comparison results of multi-category object counting. The first row is the original RGB image and the NIR false color image.}
	\label{fig:11}
\end{figure*}

Fig. \ref{fig:9} shows a scene with a high density of objects (2010 vehicles). Some crowd-counting methods, such as MCNN, CARNet, and SFCN, achieve relatively small errors in predicting the vehicle count. However, these methods exhibit overlapping predictions in the density maps for other object categories. Similarly, other single-category remote sensing object counting methods like ASPNet and PSGCNet also show overlapping predictions in different channels. In the prediction results of CARNet, ASPNet, and SwinCounter (columns 3, 6, 7, and 9 in Fig. \ref{fig:9}), noticeable noise can be observed in the density maps for individual object types, leading to negative values in the sum of Gaussian kernels and indicating predicted object counts below zero. 
A similar situation is observed in Fig. \mbox{\ref{fig:10}} and Fig. \mbox{\ref{fig:11}}, where other single-category counting algorithms exhibit overlaps and noise in the predicted density maps of other object categories. In contrast, our proposed method achieves more accurate predictions for densely distributed objects in the scene. Furthermore, it can predict the counts of objects that are sparsely distributed across different categories. Importantly, our method do not show noticeable overlaps or noise in the predicted density maps. This validates the effectiveness of proposed spatial contrast loss in suppressing feature representations between different density map channels and effectively modeling the mapping relationship between predicted density map channels with object categories.

\begin{table}[!ht]
	\centering
	\caption{Ablation studies on the effect of Multi-spectrum fusion. \textbf{Font Bold} Indicates the Best Performance.}
     \setlength{\tabcolsep}{5.5mm}{
	\begin{tabularx}{\linewidth}{cc|cc}
		\Xhline{1.2pt}		
		\multicolumn{2}{c|}{Component} & \multirow {2}{*}{$\mathrm{\overline{MSE}}$} & \multirow {2}{*}{WMSE}  \\
		\cline{1-2}
		NIR & Dual-attention & ~ & ~  \\
		\hline
  	\XSolidBrush & \XSolidBrush & 12.9113 & 52.3331 \\
		\Checkmark & \XSolidBrush & 11.7153 & 44.7001  \\
		\XSolidBrush & \Checkmark &  14.4097 & 47.1505  \\		
		\Checkmark & \Checkmark  &  \textbf{10.3441}  & \textbf{37.8798} \\								
		\Xhline{1.2pt}										
	\end{tabularx}}
	\label{table:3}
\end{table}

\subsection{Ablation Studies}
\subsubsection{Effect of Multi-spectrum fusion}
In this subsection, we conduct a series of ablation experiments to examine the impact of the NIR spectrum and the Dual-attention module on the performance of our proposed method. The results are presented in Table \ref{table:3}. When NIR features are not available, the input image undergoes processing through the backbone network and FPN, resulting in feature maps of the same size. Subsequently, dual-attention fusion is applied. By comparing the first and second rows in the table, it can be observed that solely using the NIR spectrum without any fusion mechanism resulted in a reduction of 1.1960 and 7.6330 in the \mbox{$\overline{MSE}$} and WMSE errors of the counting model, respectively.
When comparing the third row to the first row in the table, there is an increase in \mbox{$\overline{MSE}$}, but the WMSE decreases. This indicates that while the use of dual attention may slightly increase the average counting error between categories, it helps reduce the WMSE when considering the weights of different categories. Furthermore, by using the dual-attention module to fuse the NIR and RGB features, the counting performance of the model is further improved, resulting in a reduction of 1.3712 and 6.8203 in $\mathrm{\overline{MSE}}$ and WMSE errors, and achieving even better performance. The reason for the performance drop is that NIR features inherently offer richer discriminative information. The use of dual attention facilitates a more effective fusion of RGB and NIR data, enabling a better integration of these complementary modalities.

\begin{figure}[!ht]
	\centering
	\includegraphics[scale=0.50]{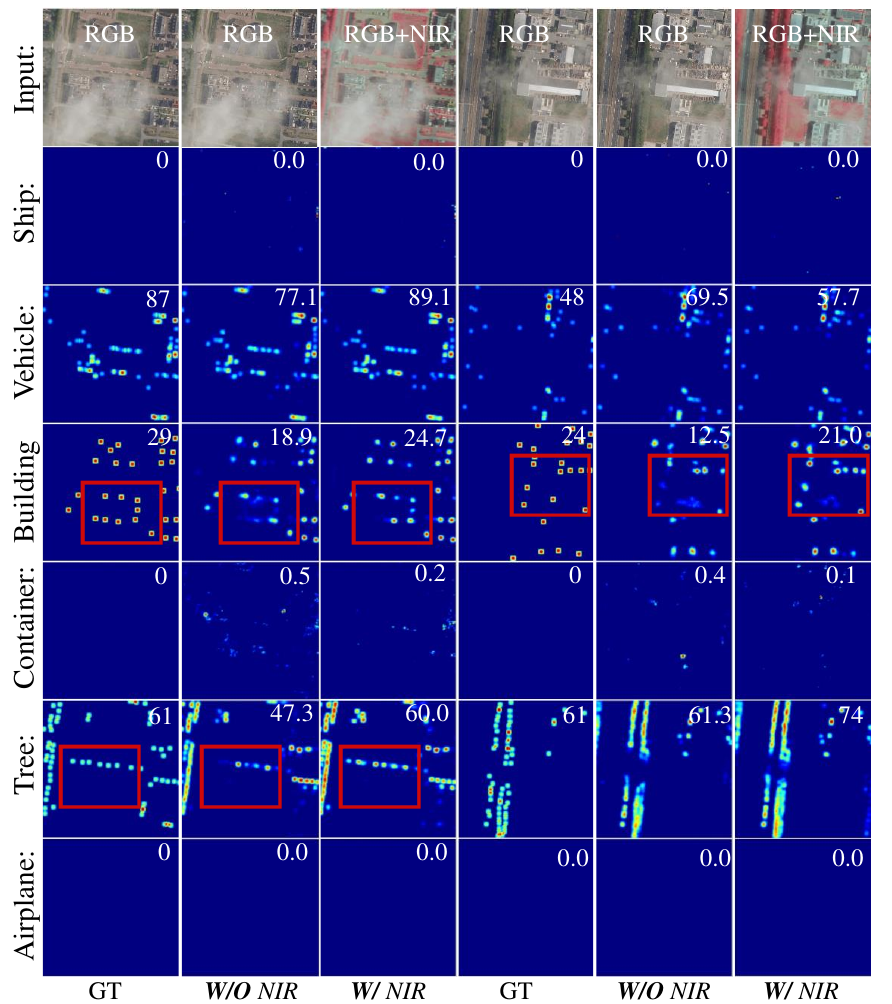}
	\caption{Comparison of visualization results before and after using NIR.}
	\label{fig:12}
 % \vspace{-0.4cm}
\end{figure}

In Fig. \ref{fig:12}, we select scenes that are affected by weather conditions and compare the counting visualization results between using NIR-RGB fusion and using only RGB input. For the objects occluded by weather conditions, as indicated by the red boxes in the figure, the use of NIR information still enables the detection and recognition of these objects.
These ablation experiments provide evidence that the inclusion of the NIR spectrum as an input can enhance the object counting performance in aerial scenes. Moreover, the introduced dual-attention module demonstrates its effectiveness in integrating the NIR and RGB feature information.

% \vspace{-0.4cm}
\begin{table}[!ht]
	\centering
	\caption{Ablation studies on the effect of Spatial Contrast Loss. \textbf{Font Bold} Indicates the Best Performance.}
    \setlength{\tabcolsep}{10mm}{
	\begin{tabularx}{\linewidth}{l|cc}
		\Xhline{1.2pt}		
		$\gamma$ & $\mathrm{\overline{MSE}}$ & WMSE  \\
		\hline
		1 & 17.0768 & 72.8902  \\
        0 & 12.3179 & 45.7567  \\
		0.1 & 16.0974 & 61.0165  \\	
        0.01 & 14.6669 & 73.2437  \\
        0.001 & 11.1826 & 39.1082  \\
		0.0001 & \textbf{10.3441} & \textbf{37.8798} \\		
        0.00001 & 10.3746 & 38.0919  \\
		\Xhline{1.2pt}										
	\end{tabularx}}
	\label{table:4}
\end{table}

\subsubsection{Effect of Spatial Contrast Loss}
To investigate the impact of the proposed spatial contrast loss $\mathcal{L}_{\mathcal{S}}$ and the hyper-parameter $\gamma$ on the counting performance of the model, we conduct a series of ablation experiments in this part. The objective is to determine the optimal value for $\gamma$ and assess the effectiveness of the spatial contrast loss. While maintaining other experimental conditions constant, we vary the value of $\gamma$ and evaluate its effect on the counting errors, as shown in Table \ref{table:4}. From the results in the table, it can be observed that as $\gamma$ decreases from 1 to 0.0001, the model achieves smaller object counting errors, with the lowest error obtained at $\gamma = 0.0001$. Therefore, we select $\gamma = 0.0001$ as the hyper-parameter for model training. Moreover, it is evident that models utilizing the spatial contrast loss significantly reduce both the ($\mathrm{\overline{MSE}}$) and WMSE errors compared to models without the spatial contrast loss ($\gamma = 0$ in Table \ref{table:4}).

\begin{figure}[!ht]
	\centering
	\includegraphics[scale=0.50]{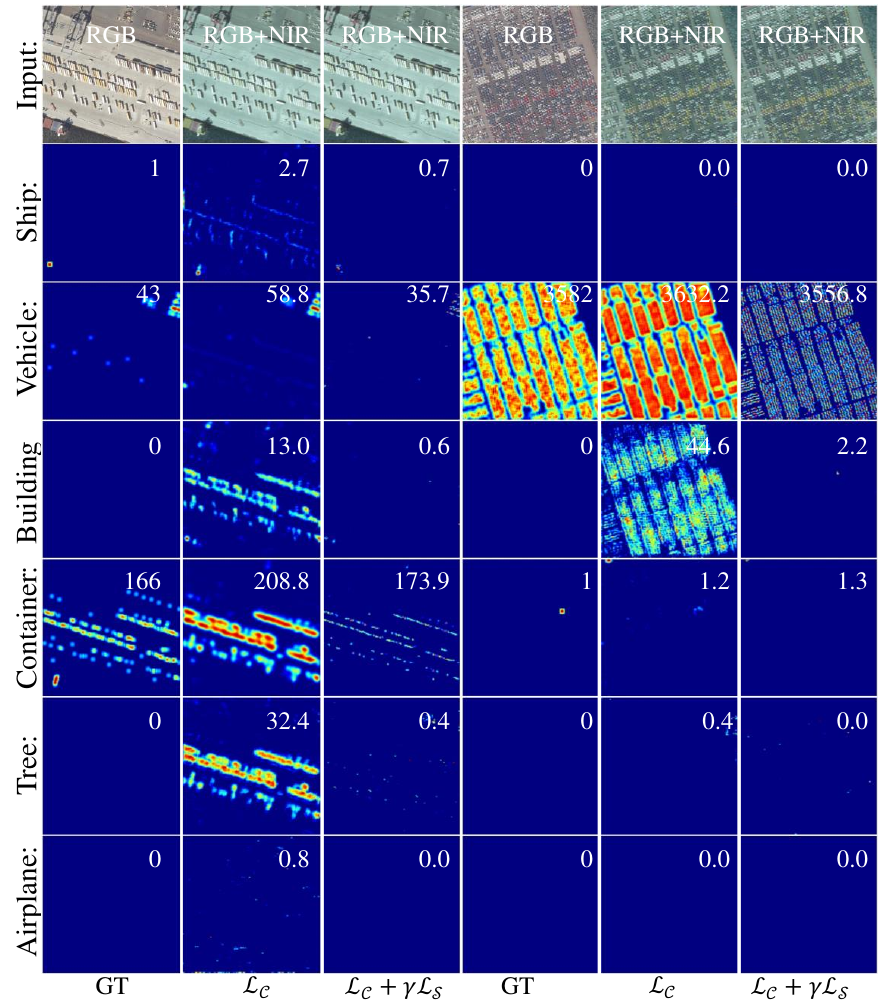}
	\caption{Comparison of visualization results before and after using $\mathcal{L}_{\mathcal{S}}$.}
	\label{fig:13}
\end{figure}

Furthermore, Fig. \ref{fig:13} provides visualizations of the density maps, supporting the same conclusion. In the visualization results, models trained solely with $\mathcal{L}_{\mathcal{C}}$ as the loss function exhibit overlaps in the other channels of the predicted density map. In contrast, models trained with $\mathcal{L}_{\mathcal{C}} + \gamma  \mathcal{L}_{\mathcal{S}}$ ($\gamma = 0.0001$) effectively reduce these overlaps, resulting in predicted density maps that closely resemble the ground truth (GT). 
Due to the mutual interference of predicted density maps in the shared feature space, such overlapping predictions may occur. However, by introducing the spatial contrast loss, the spatial similarity between channels is reduced, mitigating the occurrence of overlapping predictions between channels.
The aforementioned experimental results validate the efficacy of the proposed spatial contrast loss for modeling the mapping relationship between the predicted density maps and the GT.
% \vspace{-0.2cm}
\subsubsection{Multi-scale Feature Fusion.}
\begin{table}[!ht]
	\centering
	\caption{Ablation studies on feature fusion at different scales in FPN, where $N$ denotes the number of fused features.}
    \setlength{\tabcolsep}{3.2mm}{
	\begin{tabularx}{\linewidth}{l|llll}
		\Xhline{1.2pt}		
		$N$ & $\mathrm{\overline{MSE}}$ & WMSE & FLOPs (G) & Params. (M)  \\
		\hline
		2 & 17.0768 & 72.8902  & 79.6 & 15.5 \\
        3 & 10.3441 & 37.8798  & 80.4 & 59.9 \\
		4 & 11.0348 & 37.6354 & 90.1  & 213.4\\	
		5 & 10.2356 & 35.1487 & 121.7 & 851.7 \\		
		\Xhline{1.2pt}										
	\end{tabularx}}
	\label{table:5}
\end{table}
In this part, we explore the impact of the different number of multi-scale feature maps from the backbone and fusing them using FPN on the performance of the model. As shown in Table \mbox{\ref{table:5}}, while using more scales in the FPN leads to a slight improvement in performance, it also results in a drastic increase in the number of model parameters and FLOPs. Therefore, we ultimately chose $N=3$ to balance computational complexity and model performance, which involves fusing features from three different scales in the FPN.

\begin{table}[!ht]
	\centering
	\caption{Counting results of MCC as the baseline model on MOC6 vs. counting results on MOC14.}
    \setlength{\tabcolsep}{2.6mm}{
    \renewcommand{\arraystretch}{1.0}
	\begin{tabularx}{\linewidth}{lll|lll}
		\Xhline{1.2pt}		
		MOC-6 &    MAE  & RMSE     & MOC-14 & MAE & RMSE \\
		\hline
		Tree & 15.2229 & 32.0804  & Tree  & 16.6072   & 35.3590  \\   \hline  
		Container & 1.9558 & 7.7771  & Container  & 1.6381   & 6.2787  \\   \hline  
		Airplane & 0.0692 & 0.4780  & Airplane  &  0.0615 & 0.4768 \\    \hline  
        \multirow{2}{*}{Ship}  & \multirow{2}{*}{1.0052} & \multirow{2}{*}{4.9582} & Boat  & 0.7884   & 4.8291  \\
		~ & ~ & ~  & Vessel  & 0.1629   & 1.4042 \\ \hline
		\multirow{2}{*}{Vehicle}  & \multirow{2}{*}{4.1862} & \multirow{2}{*}{10.7381} & Car  & 5.6273   & 14.4216  \\
		~ & ~ & ~  & Truck  & 1.6133   & 5.1053  \\  \hline  
        \multirow{5}{*}{Building}  & \multirow{5}{*}{2.9002} & \multirow{5}{*}{6.2173} & House  & 2.8895   & 6.9919  \\
		~ & ~ & ~  & Industrial  & 0.9073   & 2.7487  \\ 
		~ & ~ & ~  & Mansion  & 1.7766   & 4.9877  \\ 
		~ & ~ & ~  & Stadium  & 0.1463   & 0.8543  \\ 
		~ & ~ & ~  & Others  & 0.2107   & 1.3543  \\ \hdashline
		\multicolumn{1}{l}{$\mathrm{\overline{MSE}}$}& \multicolumn{2}{c|}{11.0374 }  &  \multicolumn{3}{c}{7.0676}   \\ \hline     
		\multicolumn{1}{l}{WMSE}& \multicolumn{2}{c|}{42.0919}  &  \multicolumn{3}{c}{49.1304}   \\    
	
		\Xhline{1.2pt}										
	\end{tabularx}
 }
	\label{table:6}
\end{table}
\subsection{Fine-grained Object Counting on MOC14}
In this subsection, we perform object counting for the fine-grained categories in MOC14. Here, we use MCC as the baseline model, only modifying the channel number in its output density maps.  As shown in Table \mbox{\ref{table:6}}, the results on the left side are from MOC6, while those on the right side are the counting results from MOC14. In scenarios with more categories, MAE and RMSE may not adequately reflect the generalization capacity of the model for long-tailed distributions. Therefore, although MCC outperforms on MAE and RMSE for several categories in the MOC14 dataset, when considering overall WMSE, MCC excels on MOC6. This indicates that compared to the counting results on MOC6, the performance of the baseline model in MOC14 is evidently less satisfactory. Besides the impact of category long-tail distribution on the model, the spatial resolution of the constructed dataset is 0.25 pixel / m, resulting in small object sizes in the corresponding images. Additionally, there is minimal inter-class variation among the fine-grained object categories. Hence, discerning and counting these fine-grained object categories in the images is a quite challenging task.

\section{Conclusion}
This paper introduces a new vision task in aerial scenes, Multi-category Object Counting (MOC). To handle this task, we construct a large-scale dataset (NWPU-MOC) that contains 3, 416 aerial scenes, each including RGB and NIR images, and annotate the objects with 14 categories. Based on the NWPU-MOC, a multi-category, multi-spectral object counting framework is proposed, in which a spatial contrast loss is designed for modeling the mapping between density maps and object categories. In addition, a new evaluation metric is presented to reasonably assess the performance of the MOC algorithm.
In the future, we will continue to focus on this task, aiming to achieve even finer-grained counting within the MOC-14 dataset. Additionally, we will explore mechanisms for fusing multi-spectral features and analyzing spatial relationships among different objects.

\bibliographystyle{IEEEbib}
\bibliography{refs}

\vfill

\end{document}